\documentclass[journal]{IEEEtran}

\usepackage[linesnumbered,ruled]{algorithm2e}
\usepackage{graphicx,amssymb,mathrsfs,amsmath,pifont,amscd,latexsym,amsfonts,amsthm,colortbl}
\usepackage{graphics}
\usepackage{epstopdf}
\usepackage{float}
\usepackage[scriptsize]{subfigure}
\usepackage{epsfig}
\usepackage[font=scriptsize]{caption}
\usepackage{amsfonts}
\usepackage{setspace}
\usepackage{xcolor}
\usepackage{multirow}
\usepackage{soul}
\newcommand{\norm}[1]{\lVert#1\rVert}
\usepackage{algpseudocode}
\usepackage{cite}

\DeclareMathOperator{\F}{F}
\DeclareMathOperator{\T}{T}
\begin{document}

\title{An Augmented Linear Mixing Model to Address Spectral Variability for Hyperspectral Unmixing}

\author{Danfeng Hong,~\IEEEmembership{Student Member,~IEEE,}
        Naoto Yokoya,~\IEEEmembership{Member,~IEEE,}
        Jocelyn Chanussot,~\IEEEmembership{Fellow,~IEEE,}
        and~Xiao~Xiang~Zhu,~\IEEEmembership{Senior Member,~IEEE}
\thanks{This paper is the extended version of \cite{hong2017learning} accepted by ICIP2017 . This work was supported by funding from the European Research Council (ERC) under the European Union's Horizon 2020 research and innovation program (grant agreement No [ERC-2016-StG-714087], Acronym: \textit{So2Sat}), Helmholtz Association under the framework of the Young Investigators Group ''SiPEO'' (VH-NG-1018, www.sipeo.bgu.tum.de), and the Bavarian Academy of Sciences and Humanities in the framework of Junges Kolleg. This work of N. Yokoya was supported by the Japan Society for the Promotion of Science (KAKENHI 18K18067). This work of J. Chanussot was partially supported by ANR ASTRID (project APHYPIS) under grant ANR-16-ASTR-0027-01.}
\thanks{D. Hong and X. Zhu are with the Remote Sensing Technology Institute (IMF), German Aerospace Center (DLR), 82234 Wessling, Germany, and Signal Processing in Earth Observation (SiPEO), Technical University of Munich (TUM), 80333 Munich, Germany. (e-mail: danfeng.hong@dlr.de; xiaoxiang.zhu@dlr.de)}
\thanks{N. Yokoya is with the RIKEN Center for Advanced Intelligence Project, RIKEN, 103-0027 Tokyo, Japan. (e-mail: naoto.yokoya@riken.jp)}
\thanks{J. Chanussot is with Univ. Grenoble Alpes, CNRS, Grenoble INP, GIPSA-lab, F-38000 Grenoble, France, also with the Faculty of Electrical and Computer Engineering, University of Iceland, Reykjavik 101, Iceland. (e-mail: jocelyn@hi.is)}
}

\markboth{IEEE Transactions on Image Processing,~Vol.~XX, No.~XX, ~XXXX,~2018}%
{Shell \MakeLowercase{\textit{et al.}}: An Augmented Linear Mixing Model to Address Spectral Variability for Hyperspectral Unmixing}
\maketitle

\begin{abstract}
\textcolor{blue}{This is the pre-acceptance version, to read the final version please go to IEEE Transactions on Image Processing on IEEE Xplore.} Hyperspectral imagery collected from airborne or satellite sources inevitably suffers from spectral variability, making it difficult for spectral unmixing to accurately estimate abundance maps. The classical unmixing model, the linear mixing model (LMM), generally fails to handle this sticky issue effectively. To this end, we propose a novel spectral mixture model, called the augmented linear mixing model (ALMM), to address spectral variability by applying a data-driven learning strategy in inverse problems of hyperspectral unmixing. The proposed approach models the main spectral variability (i.e., scaling factors) generated by variations in illumination or typography separately by means of the endmember dictionary. It then models other spectral variabilities caused by environmental conditions (e.g., local temperature and humidity, atmospheric effects) and instrumental configurations (e.g., sensor noise), as well as material nonlinear mixing effects, by introducing a spectral variability dictionary. To effectively run the data-driven learning strategy, we also propose a reasonable prior knowledge for the spectral variability dictionary, whose atoms are assumed to be low-coherent with spectral signatures of endmembers, which leads to a well-known low-coherence dictionary learning problem. Thus, a dictionary learning technique is embedded in the framework of spectral unmixing so that the algorithm can learn the spectral variability dictionary and estimate the abundance maps simultaneously. Extensive experiments on synthetic and real datasets are performed to demonstrate the superiority and effectiveness of the proposed method in comparison with previous state-of-the-art methods.
\end{abstract}

\begin{IEEEkeywords}
Alternating direction method of multipliers, low-coherent dictionary learning, remote sensing, spectral unmixing, spectral variability.
\end{IEEEkeywords}

\graphicspath{{figures/}}

\section{Introduction}
\IEEEPARstart{W}{ith} the rapid development of imaging spectrometers, considerable attention has been paid to spectral-based data processing and analysis, including dimensionality reduction \cite{ma2016spatial}\cite{hong2016local}\cite{Hong2017DR}, spectral unmixing \cite{pan2016fast}\cite{hong2018SULoRA}, segmentation \cite{Veganzones2014Segmentation}, classification \cite{Yokoya2015classification}\cite{hong2018joint}, and object detection and recognition \cite{yokoya2015object}\cite{hong2015novel}. Many pixels in hyperspectral data suffer from the effect of material mixtures due to a lower spatial resolution than that of color or multispectral imaging and so on. Mixed pixels inevitably degrade the performance of high-level data analysis. Therefore, spectral unmixing has been gaining importance for hyperspectral image analysis. Hyperspectral unmixing is a procedure that decomposes the measured pixel spectrum of hyperspectral data into a collection of constituent spectral signatures (or \textit{endmembers}) and a set of corresponding fractional abundances. Hyperspectral unmixing techniques have been widely used for a variety of applications \cite{SU_survey}, such as mineral mapping\cite{rogge2006iterative} and land-cover change detection \cite{adams1995classification}.

The linear mixing model (LMM) is a simple but effective model that is extensively used for spectral unmixing. However, two main factors, nonlinearity and spectral variability, still hinder the LMM's ability to yield high performance. In hyperspectral imaging, nonlinearity - i.e., nonlinearly mixed spectral signatures - is the result of multiple scattering and intimate mixing. Spectral variability refers to a variation of a spectral signature for a given material, due to illumination conditions and topography, atmospheric effects, or even the intrinsic variability of the material \cite{Somers2011SVreview} \cite{Zare2014SVreview}. Quite recently, considerable attention has been paid to dealing with spectral variability in hyperspectral unmixing \cite{Zare2014SVreview,LucasWHISPERS16,halimi2016hyperspectral,uezato2016incorporating}. Variations of spectral signatures for a material can result in significant errors in hyperspectral unmixing.

In the literature, several theories have been proposed to model spectral variability. In \cite{Eches2010Bayesian} and \cite{Du2014Beta}, the normal compositional model and the beta compositional model were designed by assuming that spectral variability follows a given probability distribution. Fu et al. proposed a spectral-library-based spectral unmixing approach, called the dictionary-adjusted nonconvex sparsity-encouraging regression (DANSER), to model the mismatch between the spectral library and the observed spectral signatures \cite{Fu2016DANSER}. This kind of mismatch can be also treated as spectral variability in general. Obviously, the spectral variability in a certain scene can hardly be modeled by giving an explicit distribution in reality. Thouvenin \textit{et al.} \cite{Thouvenin2016PLMM} indicated that spectral variability can be represented using a perturbed linear mixing model (PLMM), where the variability is explained by an additive perturbation term for each endmember. One drawback of this model is a lack of physical meaning. For instance, as a principal spectral variability, scaling factors should be coherent with endmember spectral signatures, while other variabilities are often incoherent with endmember spectral signatures. Intuitively, such attributed spectral variability can not be represented by an additional term. In contrast, an interesting approach, called an extended linear mixing mode (ELMM) , has been proposed in \cite{Veganzones2014ELMM} \cite{Drumetz2016ELMM} . This work mainly focuses on modeling the scaling factors on the endmembers, but is a slight deficiency in that other spectral variabilities cannot be considered correspondingly. Only taking the scaling factors into account is incomplete due to those innegligible spectral variabilities (e.g. atmospheric effects or nonlinear spectral mixing) that are restrictively represented only using scaling factors. Figs. \ref{fig:exampleSVa} and \ref{fig:explainEX1} show the intuitive examples to clarify the significance of considering other spectral variabilities.

To address the limitations of the PLMM and the ELMM, the purpose of this paper is to model the scaling factors and other spectral variability simultaneously, according to their distinctive properties. More specifically, our contributions can be summarized as follows:
\begin{itemize}
\item We propose a novel spectral mixture model, called an augmented linear mixing model (ALMM), where scaling factors are modeled by the endmember dictionary and an additional dictionary is introduced to model the rest of spectral variabilities simultaneously;
\item A data-driven dictionary learning method is explored in the proposed framework of spectral unmixing in which a statistical prior is given, specifying that the spectral variability (except for scaling factors) be low-coherent with endmember spectral signatures;
\item An optimization algorithm based on the alternating direction method of multipliers (ADMM) is designed to solve the proposed model.
\end{itemize}

The remainder of this paper is organized as follows. Section \uppercase\expandafter{\romannumeral2} describes the classical LMM and its variations, particularly analyzing their advantages and disadvantages. In Section \uppercase\expandafter{\romannumeral3}, we elaborate on our motivation and propose the methodology for the novel spectral mixture model (ALMM) and the corresponding optimization algorithm. Section \uppercase\expandafter{\romannumeral4} presents the experimental results using three different datasets and discusses the qualitative and quantitative analysis. Finally, Section \uppercase\expandafter{\romannumeral5} concludes with a summary.

\section{The Linear Mixing Model and Its Variations}
\label{sec:format}
\noindent In this section, we introduce the LMM and discuss its variations, the ELMM and the PLMM. Their respective motivations to address spectral variability are presented and analyzed in detail, with a focus on their advantages and disadvantages.
\subsection{Linear Mixing Model}
\label{ssec:subhead}
Let $\mathbf{Y}=\lbrack \mathbf{y}_{1},...,\mathbf{y}_{k},...,\mathbf{y}_{N}\rbrack \in\mathbb{R}^{D\times N}$ be an observed hyperspectral image with $D$ bands and $N$ pixels,
and $\mathbf{A}=\lbrack \mathbf{a}_{1},...,\mathbf{a}_{P}\rbrack \in\mathbb{R}^{D\times P}$ be the endmember matrix (or dictionary), where $P$ is the number of endmembers. $\mathbf{X}=\lbrack \mathbf{x}_{1},...,\mathbf{x}_{k},...,\mathbf{x}_{N}\rbrack \in\mathbb{R}^{P\times N}$ is the abundance map, with each column vector representing the abundance vector at each pixel. $\mathbf{R}=\lbrack \mathbf{r}_{1},...,\mathbf{r}_{k},...,\mathbf{r}_{N}\rbrack \in\mathbb{R}^{D\times N}$ is the corresponding residual matrix containing the additive noise and other errors.

With these notations, the LMM can be modeled, based on pixel-wise $\mathbf{y}_{k}\in\mathbb{R}^{D\times 1}$, as
\begin{equation}
\label{eq1}
\begin{aligned}
	  \mathbf{y}_{k} =\mathbf{A}\mathbf{x}_{k} + \mathbf{r}_{k},
\end{aligned}
\end{equation}
with the two reasonable constraints adapting to reality \cite{Heinz2001FCLSU} as follows: 1) the abundance non-negative constraint (ANC), namely $\mathbf{x}_{k}\succeq 0$; and 2) the abundance sum-to-one constraint (ASC), namely $\mathbf{1}^{T}_P\mathbf{x}_{k}=1$ ($\mathbf{1}_P=[1,1,...,1]^T \in \mathbb{R}^P$).
Considering all pixels, a compact matrix form for the LMM can be written as
\begin{equation}
\label{eq2}
\begin{aligned}
	  \mathbf{Y} =\mathbf{A}\mathbf{X} + \mathbf{R}.
\end{aligned}
\end{equation}

\begin{figure}[!t]
	  \centering
		\subfigure[an example for spectral variability]{
			\includegraphics[width=4cm,height=3cm]{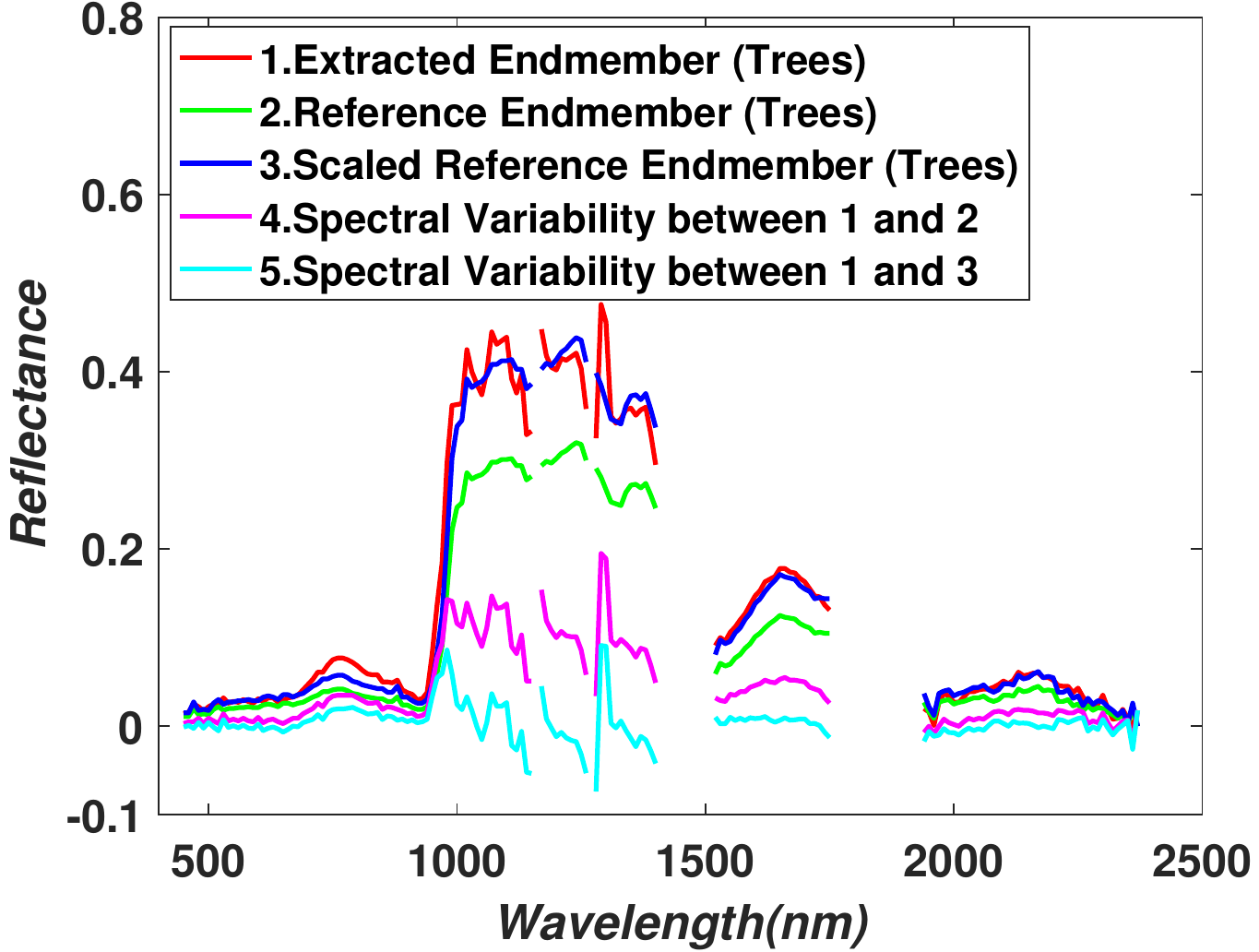}
        \label{fig:exampleSVa}
		}
		\subfigure[spectral variability distribution]{
			\includegraphics[width=4cm,height=3cm]{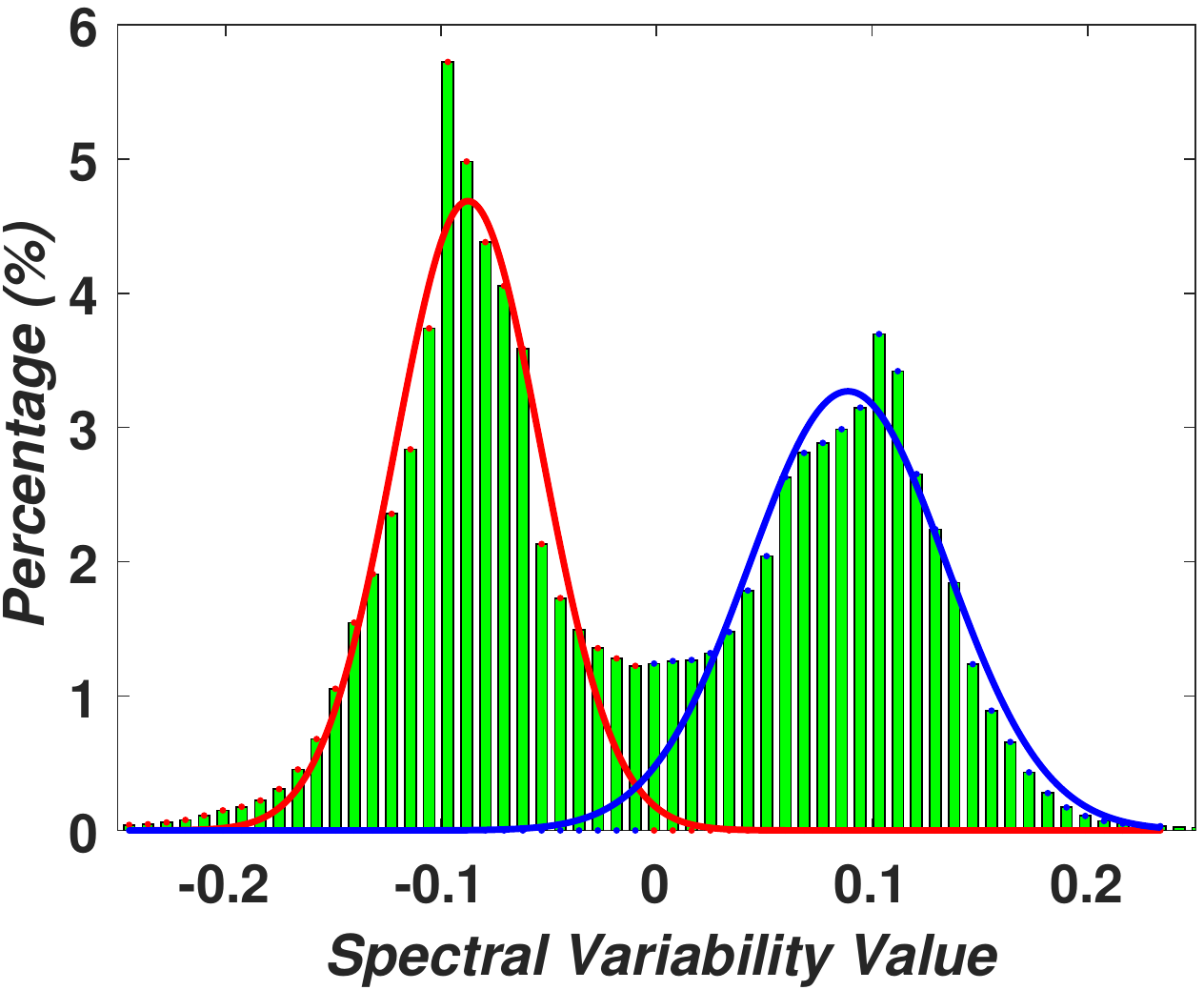}
        \label{fig:exampleSVb}
		}

         \caption{An explicit example to clarify the spectral variability. (a): The line (red) 1 denotes the endmember of the trees extracted using VCA from the Urban scene (see Section IV), while the line (green) 2 is the corresponding reference endmember (Trees) by referring to \cite{Liu2011SU} and \cite{Wang2015UrbanTIP}. The line (blue) 3 is estimated by multiplying a scaling factor on the line 2. The line 4 (or 5) illustrates the differences between 1 and 2 (or 3) in order to clarify the existence of other spectral variabilities besides scaling factors. (b) gives a statistical distribution of spectral variability in Urban scene that it is not a simple Gaussian distribution rather than more like a more complex Gaussian mixture distribution (please refer to the Section II.D for more details).}
\label{fig:exampleSV}
\end{figure}

The LMM is an approximation of reality and the linearity assumption can hold in most real cases, but it is limited to handle the problem with spectral variability. Further, these changes or effects, as often as not, lead to more specific spectral variabilities, such as scaling factors, offset, or complex noise. Although the LMM coupled with the spectral bundles technique has provided a consideration for spectral variability, spectral bundles rely heavily on establishing a good dictionary. Coincidentally, it is barely possible to prepare a good dictionary in a real case, resulting in the failure of the LMM against spectral variability. Fig. \ref{fig:exampleSVa} shows an example of spectral variability. We extracted the pure endmember (trees) via the vertex component analysis (VCA) \cite{BDias2005VCA} algorithm from an urban dataset (see  Section IV), which can be simply identified by the reference endmembers \cite{Zhu2014Urban}\cite{Wang2015UrbanTIP}. The differences between the two endmembers, (spectral variability) can be visually observed as shown in Fig. \ref{fig:exampleSVa} ({\it curve 4}). As shown in Fig. \ref{fig:exampleSV}, the extracted endmember ({\it curve 1}) can be better approximated by a scaled version of the reference endmember ({\it curve 3}) than directly by the reference endmember without a scaling factor ({\it curve 2}).
\subsection{Extended Linear Mixing Model}
Incorporating that estimation approach into the algorithm, Drumetz et al. \cite{Drumetz2016ELMM} proposed the ELMM to fully consider the scaling factors in order to allow a pixel-wise variation of each endmember:
\begin{equation}
\label{eq3}
\begin{aligned}
	  \mathbf{y}_{k} =\mathbf{A}\mathbf{S}_{k}\mathbf{x}_{k} + \mathbf{r}_{k},
\end{aligned}
\end{equation}
where $\mathbf{S}_{k}\in\mathbb{R}^{P\times P}$ is a diagonal matrix with the constraint that diagonal elements are nonnegative. Eq. (\ref{eq3}) can be extended to a compact matrix form:
\begin{equation}
\label{eq4}
\begin{aligned}
	  \mathbf{Y} =\mathbf{A}(\mathbf{S} \odot \mathbf{X}) + \mathbf{R},
\end{aligned}
\end{equation}
where $\mathbf{S}\in\mathbb{R}^{P\times N}$ aims at representing all scaling factors for all pixels whose $k^{th}$ column is $\mathbf{S}_{k}$. The mathematical symbol $\odot$ denotes the Schur-Hadamard (termwise) product.

Typically, Eqs. (\ref{eq3}) and (\ref{eq4}) are non-convex optimization problems, which difficultly provide the analytic solutions. In \cite{Drumetz2016ELMM}, the authors relaxed Eqs. (\ref{eq3}) and (\ref{eq4}) by employing a strategy of splitting variables, thereby obtaining the following objective function:
\begin{equation}
\label{eq5}
\begin{aligned}
	  \{\hat{\mathbf{X}},\hat{\mathbf{S}},\hat{\mathbf{\underline{A}}}\}=\mathop{arg \min \limits_{\mathbf{X},\mathbf{S},\mathbf{\underline{A}}}}\sum_{k=1}^{N}
&(\norm{\mathbf{y}_{k}-\mathbf{A}_{k}\mathbf{x}_{k}}_{2}^{2}\\
&+\lambda_{S}\norm{\mathbf{A}_{k}-\mathbf{A}_{0}\mathbf{S}_{k}}_{F}^{2})
\end{aligned}
\end{equation}
where $\mathbf{A}_{0}$ is the reference endmember matrix, $\mathbf{\underline{A}}=\{\mathbf{A}_{k}\}$ is the collection of pixel-dependent endmember matrices, and $\lambda_{S}$ denotes the penalty parameter to balance the two separated terms. Therefore, we can iteratively optimize individual variables by alternating nonnegative least squares (ANLS) \cite{Kim2008ANLS}.
\subsection{Perturbed Linear Mixing Model}
Inspired by a model proposed in \cite{Johnson2013PLMM}, \cite{Thouvenin2016PLMM} modeled spectral variability simply and flexibly through an additive perturbation information. This model, the PLMM, is formulated by
\begin{equation}
\label{eq6}
\begin{aligned}
	  \mathbf{y}_{k} =(\mathbf{A}+\mathbf{\Delta}_{k})\mathbf{x}_{k} + \mathbf{r}_{k},
\end{aligned}
\end{equation}
where $\mathbf{\Delta}_{k} \in\mathbb{R}^{D\times P}$ denotes the perturbation of the endmember matrix $\mathbf{A}$ in the $k^{th}$ pixel, whose columns are the perturbation vectors associated with each endmember in $\mathbf{A}$. The matrix form of Eq. (\ref{eq6}) can be expressed as
\begin{equation}
\label{eq7}
\begin{aligned}
	  \mathbf{Y} =\mathbf{A}\mathbf{X}+\underbrace {\lbrack \mathbf{\Delta}_{1}\mathbf{x}_{1} | ... | \mathbf{\Delta}_{k}\mathbf{x}_{k} |...| \mathbf{\Delta}_{N}\mathbf{x}_{N} \rbrack }_\mathbf{\Delta}+\mathbf{R},
\end{aligned}
\end{equation}
where $\mathbf{\Delta}$ is $\lbrack \mathbf{\Delta}_{1}\mathbf{x}_{1} | ... | \mathbf{\Delta}_{k}\mathbf{x}_{k} |...| \mathbf{\Delta}_{N}\mathbf{x}_{N} \rbrack$.
$\mathbf{X}$ can be estimated by adopting an alternating minimization strategy based on an ADMM optimization framework \cite{Almeida2013ADMM}. Readers are referred to \cite{Thouvenin2016PLMM} for more details.
\subsection{Discussion and Summary}
In summary, according to the different prior assumptions, the LMM and its variations give the corresponding spectral mixing models for unmixing respectively.  Unfortunately, in real scenarios they are not capable of effectively dealing with spectral variability (in the case of the LMM) or can only considering a special type of spectral variability (in the case of the ELMM for scaling factors). Although the PLMM tried to create a general model that incorporates spectral variabilities, the model does not consider the properties of spectral variability (e.g., the variation of illumination conditions). Furthermore, perturbation information may explain offset variability effectively but it ignores other important spectral variabilities, like scaling factors, which results in performance degradation in the unmixing process. Fig. \ref{fig:exampleSV} shows more evidence regarding the spectral variabilities. As can be clearly seen, the spectral variability {\it curve 4} generated by the difference between {\it curve 1} and {\it curve 2} can be largely explained by the scaling factor (shown in {\it curve 3}), but spectral variability other than scaling factors still remain (refer to {\it curve 5}). This is a self-evident example that demonstrates that individually considering scaling factors or perturbed information to model spectral variability is not adequate for modeling. Nevertheless, although DANSER and PLMM attempt model spectral variability in a generalized way, in \cite{Fu2016DANSER} and \cite{Thouvenin2016PLMM} they both assume that spectral variability follows a Gaussian distribution and thus is constrained using the Frobenius norm in \cite{Fu2016DANSER,Thouvenin2016PLMM}. We have to point out, however, that spectral variability does not strictly obey a Gaussian distribution in real scenarios. Direct evidence supporting this point is shown in Fig. \ref{fig:exampleSVb}, which approximately satisfies a mixed Gaussian distribution. More specifically, we selected the potential pure endmembers from the urban data based on the reference endmembers provided and then were able to calculate the spectral variabilities between the reference endmembers and the extracted endmembers using a subtraction operation. In the end, we collected all scalars from the obtained spectral variabilities and displayed them in the form of statistics, as shown in Fig. \ref{fig:exampleSVb}.
\begin{figure*}[!t]
	  \centering
		\subfigure{
			\includegraphics[width=0.95\textwidth]{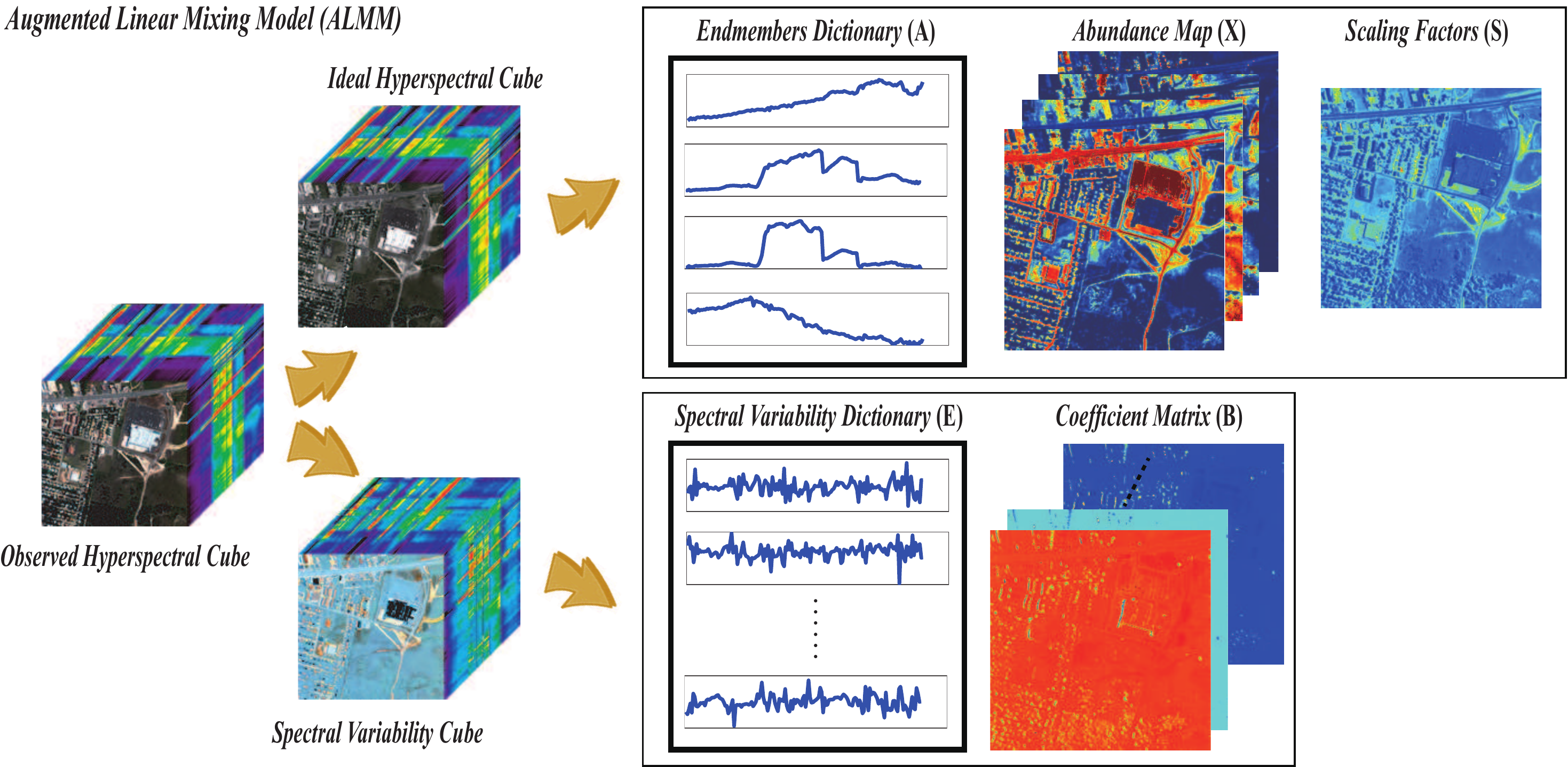}
		}
        \caption{The holistic diagram of spectral unmixing using the proposed ALMM.}
\label{fig:ALMMframework}
\end{figure*}
\section{Augmented Linear Mixing Model}
\noindent Scaling factors and other spectral variabilities are simultaneously considered in our model. Also, the reasonable prior assumptions are introduced as regularization terms into our model. Finally, an ADMM-based optimization algorithm is explored to solve the proposed model.
\subsection{Motivation}
In hyperspectral imaging, a local region in the real world, which is presented as a mixed pixel in an image, usually presents a similar scaling variability due to a similar illumination condition and topography. Considering another fact that more degrees of flexibility (e.g. endmember-wise scaling factors) are considered in the ELMM, this makes the model too ill-posed. The two facts motivated us to further slightly re-constrain the ELMM model by using a shared scaling factor on each endmember. In practice, this implementation is basically reasonable and useful as the scaling factors are strongly related to topography, which can indeed be assumed in most situations to be constant for all the endmembers of a given pixel at the scale of observation \cite{Veganzones2014ELMM}. While those small induced remaining errors that can not be represented by the shared scaling factors, should be able to be corrected with the additional degrees of liberty \footnote{For example, $\mathbf{EB}$ term which will be introduced in Eq. (\ref{eq10}).} (several corrected examples can be found in Fig. \ref{fig:exampleLSV}). Additionally, more discussions and explanations have been done in \cite{Drumetz2016ELMM} and \cite{Henrot2016dynamical}. The special case of the ELMM can be formulated as
\begin{equation}
\label{eq8}
\begin{aligned}
	  \mathbf{y}_{k} =S_{k}(\mathbf{A}\mathbf{x}_{k}) + \mathbf{r}_{k},
\end{aligned}
\end{equation}
where $S_{k}$ is a scalar in the $k^{th}$ pixel that can be simply estimated using the regression between $\mathbf{y}_{k}$ and $\mathbf{A}\mathbf{x}_{k}$. Likewise, the matrix form of Eq. (\ref{eq8}) can be written as
\begin{equation}
\label{eq9}
\begin{aligned}
	  \mathbf{Y}=\mathbf{A}\mathbf{X}\mathbf{S} + \mathbf{R},
\end{aligned}
\end{equation}
where $\mathbf{S}\in\mathbb{R}^{N\times N}$ is a diagonal matrix with its diagonal values $\mathbf{S}_{k} \succeq \mathbf{0}$.

In order to further overcome the shortcomings of the ELMM, which ignores the effects of other spectral variabilities, we extend the simplified ELMM to an augmented linear mixing model. This augmented linear mixing model, or ALMM, is expressed by
\begin{equation}
\label{eq10}
\begin{aligned}
	  \mathbf{Y}=\mathbf{A}\mathbf{X}\mathbf{S} + \mathbf{E}\mathbf{B}+\mathbf{R},
\end{aligned}
\end{equation}
where $\mathbf{E}=\lbrack \mathbf{e}_{1},...,\mathbf{e}_{m},...,\mathbf{e}_{L}\rbrack \in\mathbb{R}^{D\times L}$ denotes the spectral variability matrix (or dictionary), and $L$ is the number of basis vectors in $\mathbf{E}$. The expression $\mathbf{B}=\lbrack \mathbf{b}_{1},...,\mathbf{b}_{k},...,\mathbf{b}_{N}\rbrack \in\mathbb{R}^{L\times N}$ is the coefficient matrix corresponding to $ \mathbf{E}$.

Unlike the ELMM, where the spectral variability is modeled by endmember-wise scaling at each pixel, the ALMM represents the spectral signature by the endmember dictionary (i.e., $\mathbf{AXS}$) with pixel-wise scaling and also spectral variabilities that cannot be explained using scaling by the spectral variability term (i.e., $\mathbf{EB}$). On the other hand, unlike the PLMM, the ALMM gives an explicit physical consideration to scaling factors by inheriting concepts behind the ELMM, simultaneously modeling other variabilities by reasonable physical assumptions (see Subsection III-B for more details). Fig. \ref{fig:ALMMframework} gives the macro diagram of spectral unmixing using the proposed ALMM.

\subsection{Problem Formulation}
As introduced in Subsection III-A, the ALMM shown in Eq. (\ref{eq10}) with a non-negativity constraint can be formulated as the following constrained optimization problem:
\begin{equation}
\label{eq11}
\begin{aligned}
\mathop{arg \min}_{\mathbf{X},\mathbf{B},\mathbf{S},\mathbf{E}} &\frac{1}{2}\norm{\textbf{Y}-\textbf{AXS}-\textbf{EB}}_{\F}^{2}+\Phi(\mathbf{X})+\Psi(\mathbf{B})+\Upsilon(\mathbf{E})\\
&\mathrm{s.t.} \quad \mathbf{X}\succeq 0,\quad \mathbf{S}\succeq 0,
\end{aligned}
\end{equation}
where the intent is to estimate the variables $\mathbf{X}$, $\mathbf{S}$, $\mathbf{E}$, and $\mathbf{B}$, while $\mathbf{A}$ is given. Since Eq. (\ref{eq11}) is a typically ill-posed problem, several reasonable assumptions (or prior knowledge) should be introduced into the ALMM using regularization. Specifically, we defined three regularization functions $\Phi$, $\Psi$, and $\Upsilon$ with respect to variables $\mathbf{X}$, $\mathbf{B}$, and $\mathbf{E}$, respectively. The three regularization terms are described below.

\subsubsection{Abundance Regularization $\Phi(\mathbf{X})$}
In reality, a given spectral signature is usually composed of a limited number of materials in a hyperspectral scene, and hence the abundance regularization should be selected to be sparsity-prompting. In this paper, we applied $\norm{\textbf{X}}_{1,1}\equiv \sum_{k=1}^N\norm{\textbf{x}_k}_1$ to approximately estimate the sparsity-prompting term, which can be expressed with the penalty parameter $\alpha$ as
\begin{equation}
\label{eq12}
\begin{aligned}
       \Phi(\mathbf{X})=\alpha\norm{\mathbf{X}}_{1,1}.
\end{aligned}
\end{equation}

\subsubsection{Spectral Variability Coefficient Regularization $\Psi(\mathbf{B})$}
Spectral variability is generally generated from various factors in a given hyperspectral scene. Except for scaling factors that can be modeled well by the endmember dictionary, the rest are diverse. To achieve a reliable generalization of our model, $\mathbf{E}$ should be regularized by a Frobenius Norm parameterized by $\beta$:
\begin{equation}
\label{eq13}
\begin{aligned}
       \Psi(\mathbf{B})=\frac{\beta}{2}\norm{\mathbf{B}}_{\F}^{2}.
\end{aligned}
\end{equation}
\begin{figure*}[!t]
	  \centering
		\subfigure[]{
			\includegraphics[width=0.23\textwidth]{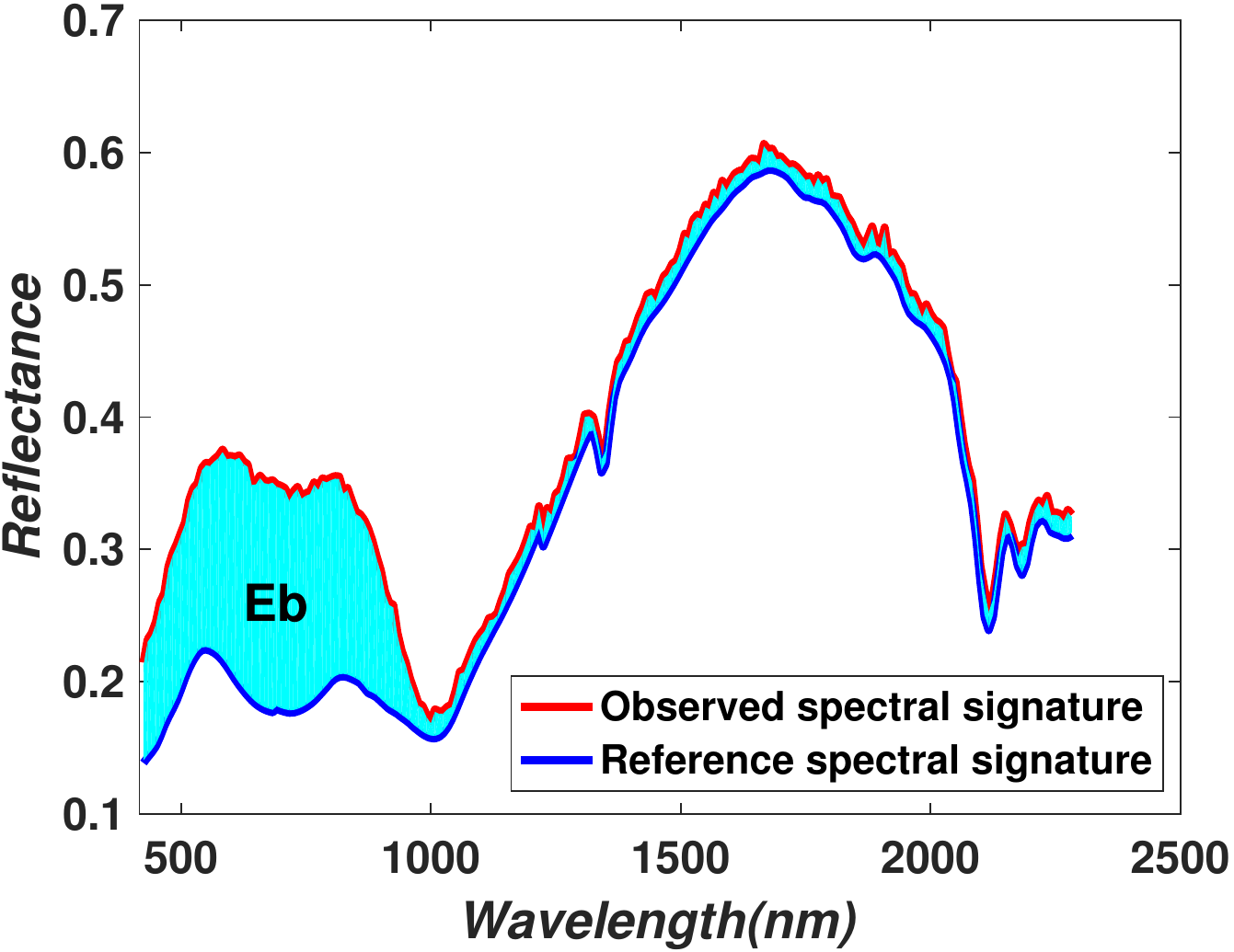}
            \label{fig:explainEX1}
		}
		\subfigure[]{
			\includegraphics[width=0.23\textwidth]{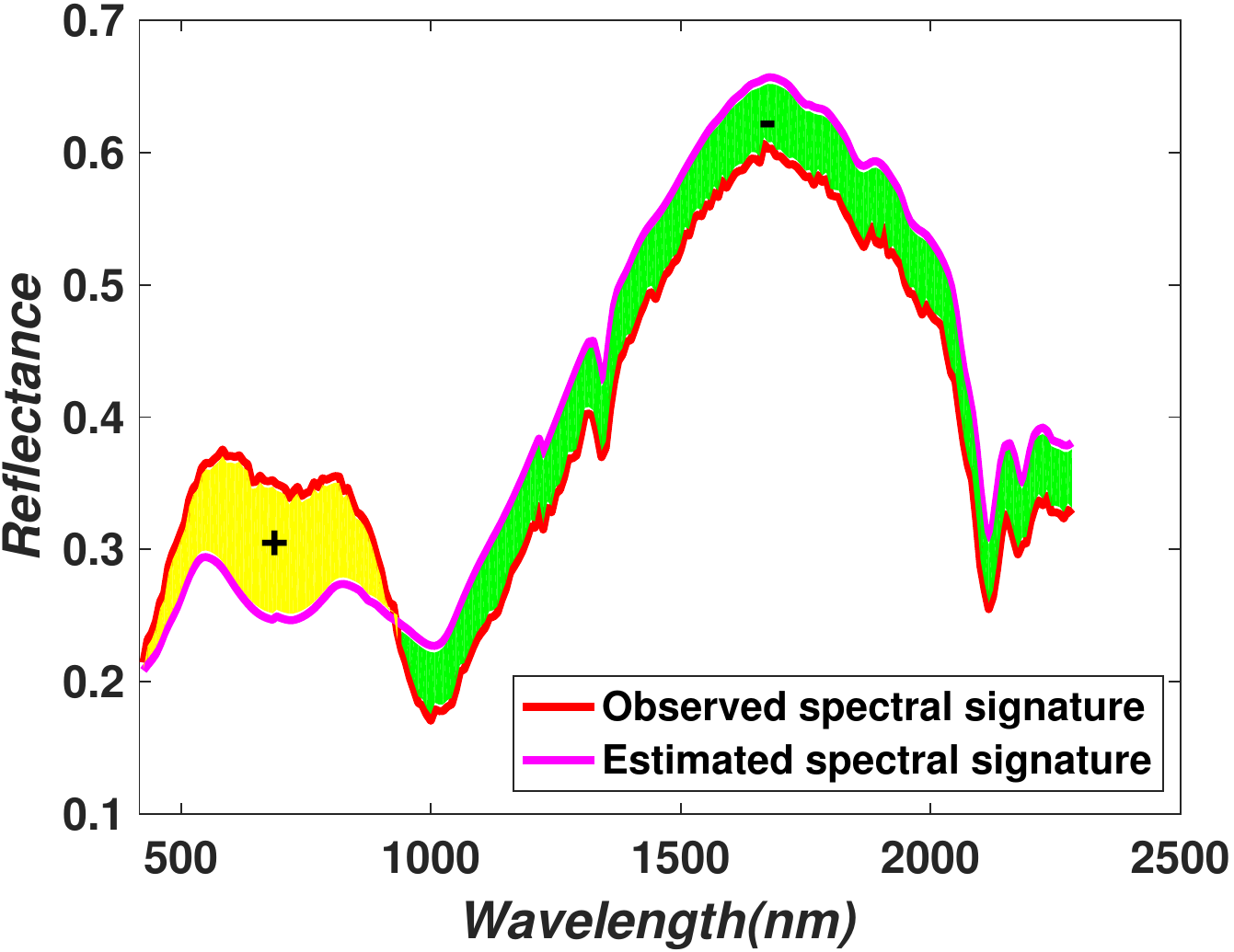}
            \label{fig:explainEX2}
		}
       \subfigure[]{
			\includegraphics[width=0.23\textwidth]{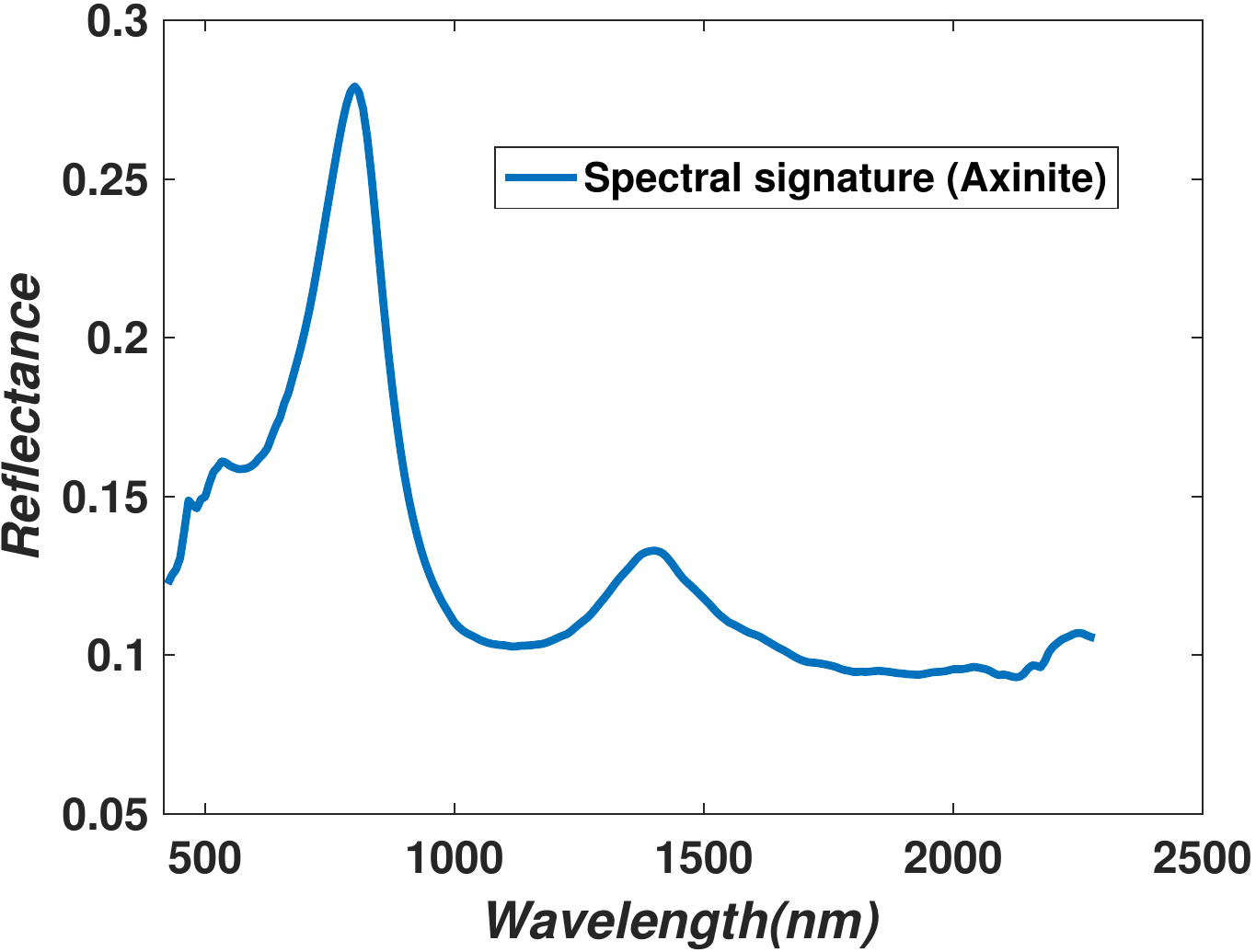}
            \label{fig:explainEX3}
		}
		\subfigure[]{
			\includegraphics[width=0.23\textwidth]{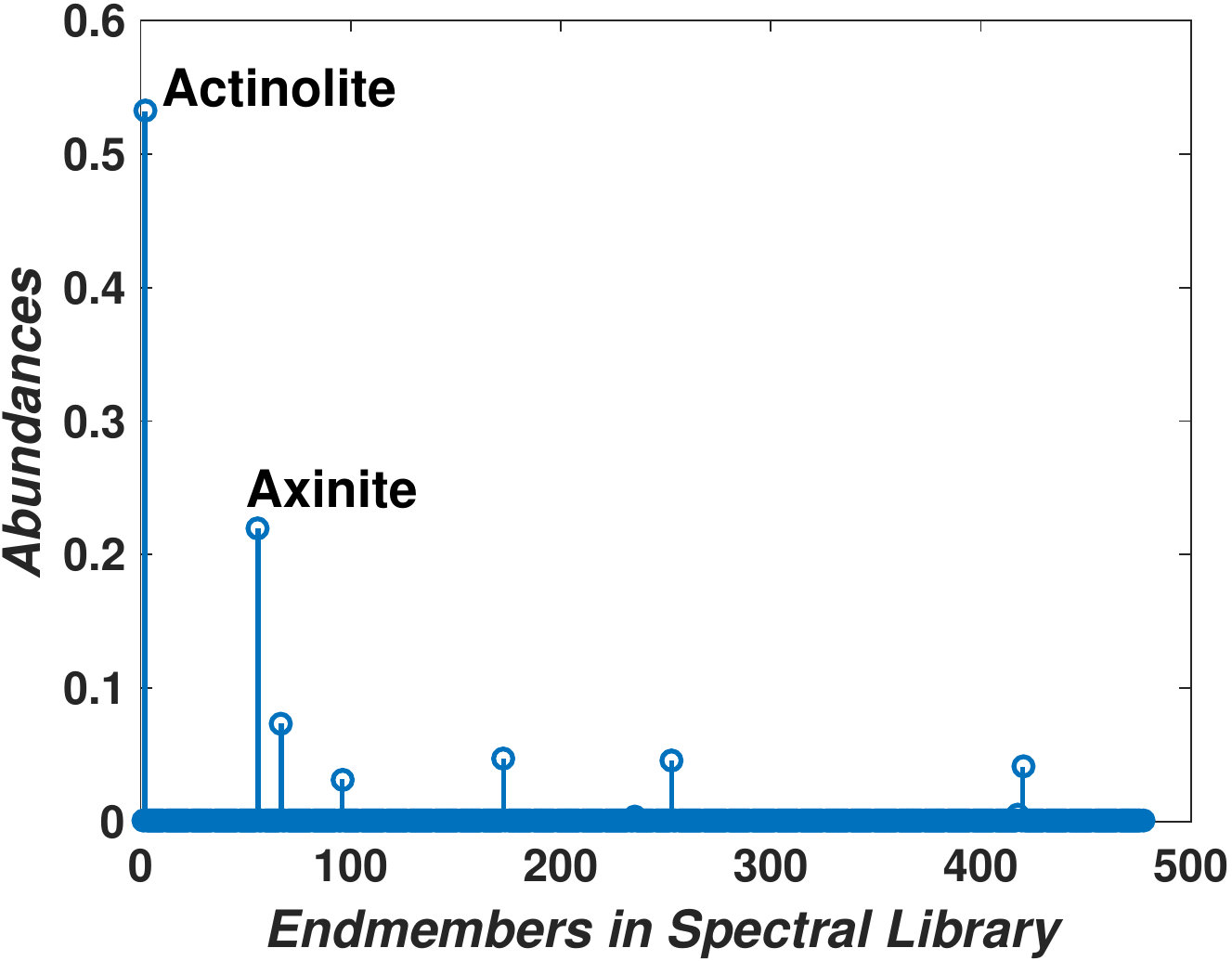}
            \label{fig:explainEX4}
		}
         \caption{An example in the real Cuprite scene to illustrate the physical meaning of $\mathbf{E}$. (a) shows the differences ($\mathbf{Eb}$) between the observed spectral signature and the real spectral signature that can not be explained by the endmember dictionary ($\mathbf{A}$), but it can be represented well by an additional spectral variability dictionary ($\mathbf{E}$). Correspondingly, if without $\mathbf{E}$, the differences (spectral variability) could be absorbed by $\mathbf{A}$ as shown in (b), leading an inaccurate estimation of abundance maps ($\mathbf{X}$). (c) gives a spectral signature of the material \textit{Axinite} and (d) shows a real case of unmixing the observed spectral signature using USGS spectral library that except the \textit{Actinolite}, the \textit{Axinite} occupies the main abundances, which can well represents the $\mathbf{E}\mathbf{b}$ in (a).}
\label{fig:explainE}
\end{figure*}
\begin{figure}[!t]
	  \centering
		\subfigure[simulated dataset]{
			\includegraphics[width=4cm,height=3cm]{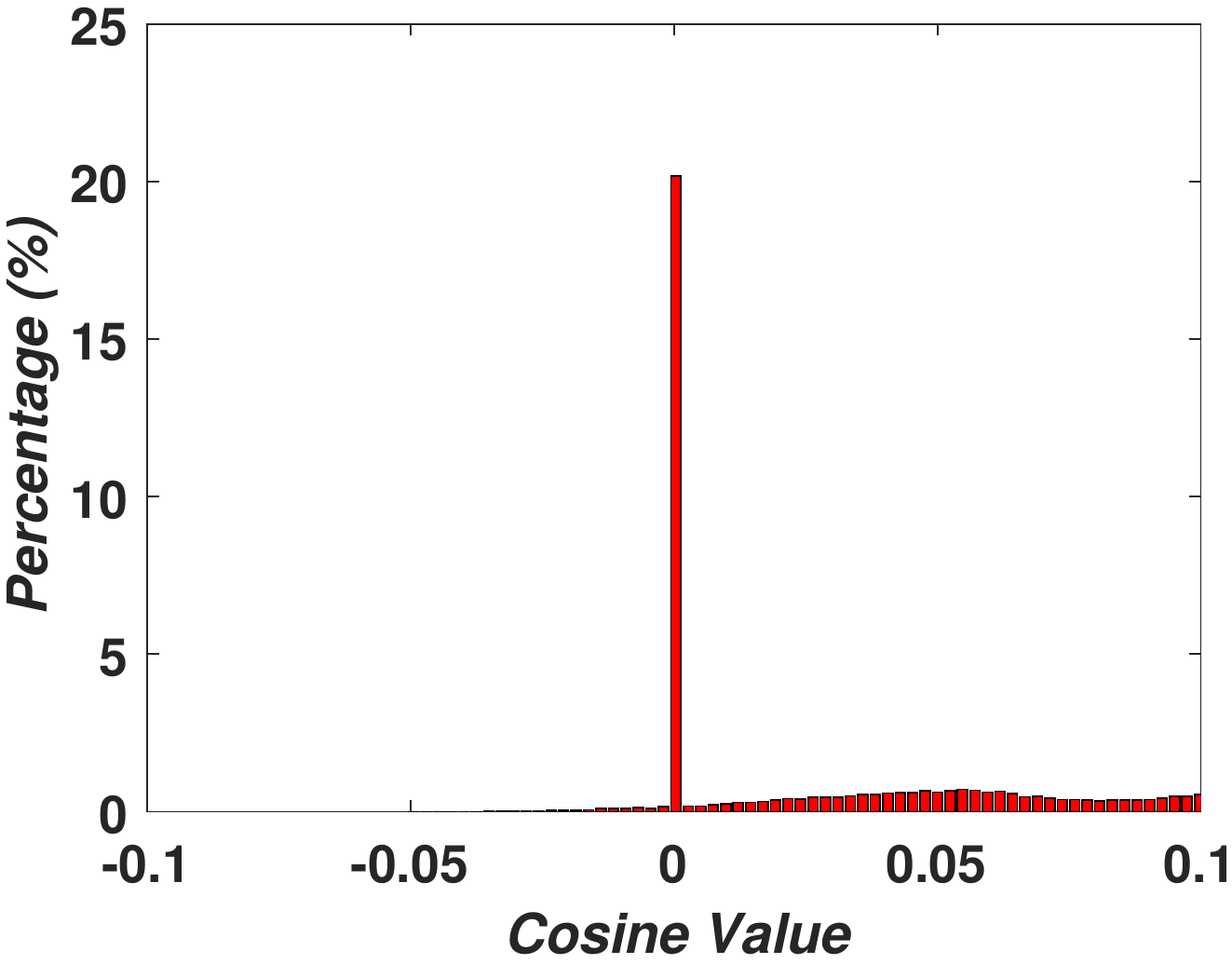}
		}
		\subfigure[real dataset (Urban)]{
			\includegraphics[width=4cm,height=3cm]{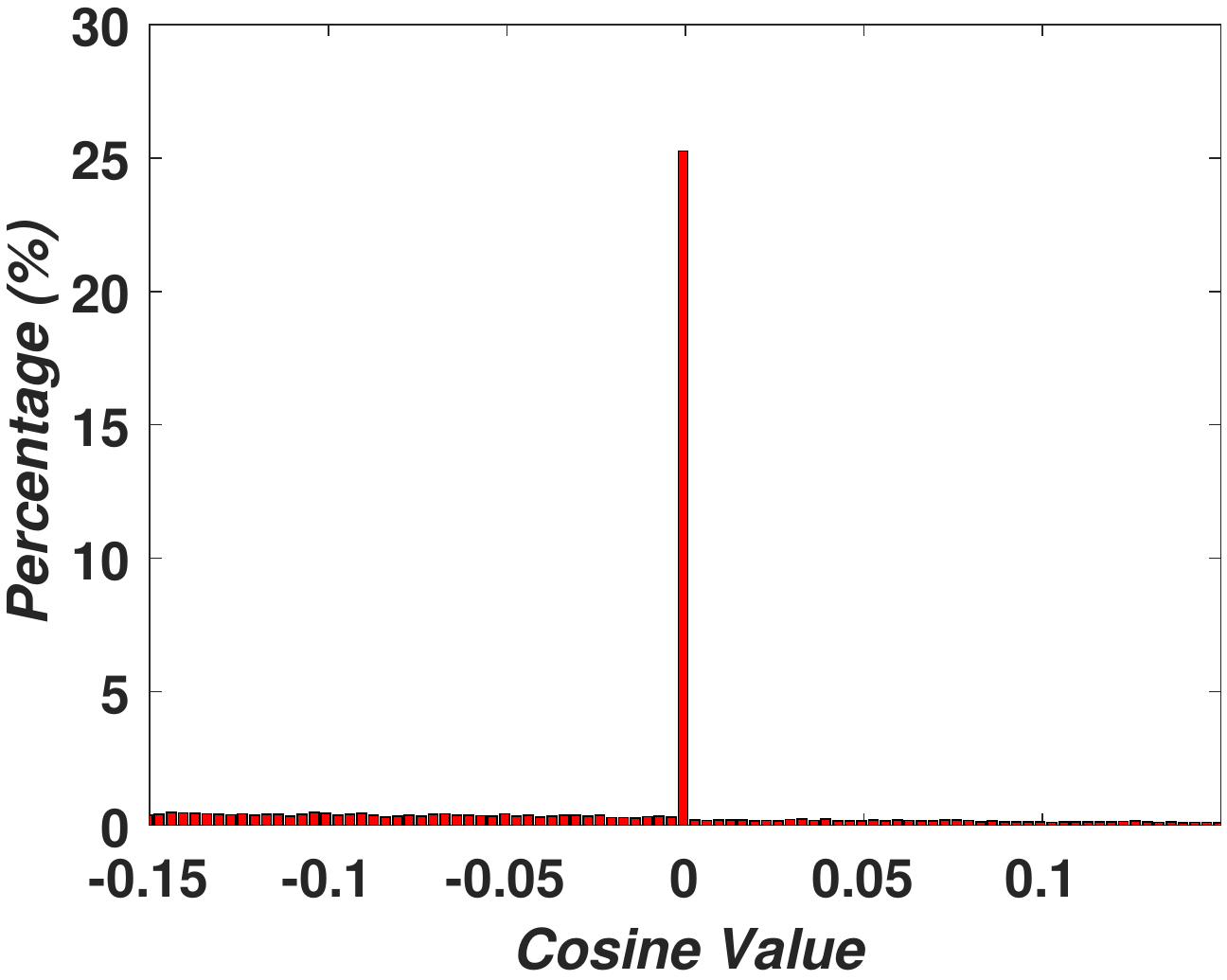}
		}

         \caption{Statistics of Cosine Value between endmembers and spectral variabilities on the first simulated dataset and real Urban scene, respectively, where the spectral variabilities are obtained by calculating the intra- and inter-class differences between the extracted endmembers and the given reference endmembers.}
\label{fig:StatisticE}
\end{figure}

\subsubsection{Spectral Variability Dictionary Regularization $\Upsilon(\mathbf{E})$}
To effectively find a better local optimal solution in our optimization problem, we acquire the variable $\mathbf{E}$ to be bounded by two prior knowledge assumptions: 1) the spectral variability dictionary ($\mathbf{E}$) should be low-coherent with the endmember dictionary ($\mathbf{A}$), formulated by $\frac{1}{2}\norm{\mathbf{A}^{T}\mathbf{E}}_{F}^{2}$. 2) $\mathbf{E}$ should possess another property, making it possible for the basis vectors of $\mathbf{E}$ to be orthogonal, since such a dictionary can adequately represent various potential spectral variabilities. This makes the second prior assumption written by $\frac{1}{2}\sum_{i=1}^{L}\sum_{j=1,j\neq i}^{L}\norm{\mathbf{e}_{i}^{T}\mathbf{e}_{j}}_{2}^{2}$. Also, the constraint $\norm{\mathbf{e}_{m}}_{2}^{2}=1 (m=1,...,L)$ should be satisfied in order to eliminate the trivial solution effectively; this second regularization term can be summarized as $\frac{1}{2}\norm{\mathbf{E}^{T}\mathbf{E}-\mathbf{I}}_{F}^{2}$ (refer to \cite{Barchiesi2013Incoherent} for more details regarding this term). The resulting expression of regularization with respect to $\mathbf{E}$ is
 \begin{equation}
\label{eq14}
\begin{aligned}
       \Upsilon(\mathbf{E})=\frac{\gamma}{2}\norm{\mathbf{A}^{\T}\mathbf{E}}_{\F}^{2}+\frac{\eta}{2}\norm{\mathbf{E}^{\T}\mathbf{E}-\mathbf{I}}_{\F}^{2},
\end{aligned}
\end{equation}
where $\gamma$ and $\eta$ are the corresponding penalty parameters.

Moreover, non-negativity constraints ($\mathbf{X}\succeq \mathbf{0}$ and $\mathbf{S}\succeq \mathbf{0}$) usually have to be considered to satisfy the physical assumption. In addition to the non-negativity constraint, the sum-to-one also plays an important role in the abundance map. However, this constraint is not considered in our original problem [Eq. (\ref{eq11})], since the variables $\mathbf{X}$ and $\mathbf{S}$ are bundled together, leading to difficultly satisfying the sum-to-one constraint for $\mathbf{X}$. In the following section, we adopt the scaled constrained least squares unmixing (SCLSU) \cite{Veganzones2014ELMM} technique to force $\mathbf{X}$ to follow the sum-to-one constraint.
\subsection{Discussion on the Physical Significance of $\mathbf{E}$}
Followed by the instruction of $\Upsilon(\mathbf{E})$ shown in Eq. (14), we attempt to further discuss and explain the physical meaning of $\mathbf{E}$, unfolded as follows:

On the one hand, although most spectral variabilities coherent with endmembers ($\mathbf{A}$) can be represented by scaling factors, yet the remaining spectral variabilities, either intra-class or inter-class, can still hurt the performance of the spectral unmixing in reality. An example is illustrated in Fig. \ref{fig:explainE} to clarify that the spectral variability can not be fully explained by the scaled endmembers. The red curve in Fig. \ref{fig:explainEX1} is the observed spectral signature (\textit{Actinolite}) extracted from the Cuprite scene and the blue one is the corresponding reference spectral signature (\textit{Actinolite}) obtained from the USGS spectral library. Obviously, the differences (or spectral variabilities) between the two curves can not be well fit by the magenta curve, as shown in \ref{fig:explainEX2}. Accordingly, we draw two points by reasoning as follows: 1) the scaled endmembers obtained by adding scaling factors on endmembers ($\mathbf{A}$) fail to fully fit the gap in-between;
2) The errors marked in cyan of Fig. \ref{fig:explainEX1} could be explained by spectral variabilities or a certain new material. We try to identify the errors by means of the USGS spectral library, generating the abundances with respect to the various materials as shown in Fig. \ref{fig:explainEX4} where there is a rather high abundance in \textit{Axinite} ranked as the second major component following \textit{Actinolite}. In our model (ALMM), we represent the errors (or spectral variabilities) by an additional spectral (variability) dictionary (e.g., $\mathbf{E}$). With the naked eye in Fig. \ref{fig:explainEX3}, the spectral signature of the \textit{Actinolite} yields a low-coherence with that of \textit{Axinite} (A statistic will be given below.).

On the other hand, the physical significance of $\mathbf{E}$ could be also explained from the perspectives of intra-class and inter-class spectral variabilities. For instance, suppose only the existence of intra-class spectral variability that can be modeled by $\mathbf{E}$, and thus the abundance maps $\mathbf{X} $ can be more accurately estimated by getting rid of the effects for the spectral variability ($\mathbf{E}\mathbf{B}$) that can not be explained by scaling factors, as shown in Figs.~\ref{fig:exampleSV} and~\ref{fig:explainE}. Without $\mathbf{E}$, the intra-class spectral variability could be absorbed by endmembers ($\mathbf{A}$), further leading an inaccurate estimation of abundance maps ($\mathbf{X}$). If $\mathbf{E}$ is considered as inter-class spectral variability dictionary, and then the term ($\mathbf{E}\mathbf{B}$) might represent the spectral signatures of certain new materials that are not discovered by the LMM (see Fig. \ref{fig:explainE} for example). The $\mathbf{E}$ used in the ALMM is therefore capable of calibrating the class-specific spectral variabilities into an unified or generalized spectral variability, which enables to simultaneously handle the intra- and inter-class variabilities. Fig. \ref{fig:StatisticE} shows a statistical evidence by collecting all cosine values between endmembers and spectral variabilities, where the cosine value is basically around 0, indicating that the spectral variability should, to a great extent, be low-coherent with the endmembers. This is basically consistent with the conclusion summarized above.
\begin{algorithm}[!t]
\footnotesize
\caption{ALMM-based pixel-wise SU}
\begin{spacing}{1}
\KwIn{$\mathbf{y}_{k},\mathbf{A},\mathbf{E},$ and parameters $\alpha, \beta, maxIter.$}
\KwOut{$\mathbf{x}_{k},S_{k},\mathbf{b}_{k}.$}
\textbf{Initialization}: $\mathbf{g}_{k}^{0}=\mathbf{h}_{k}^{0}=\mathbf{0},S_{k}^{0}=1,\mathbf{b}_{k}^{0}=\mathbf{0},\pmb{\lambda}_{k}^{0}=\pmb{\nu}_{k}^{0}=\mathbf{0},$\
$\mu^{0}=1e-3, \mu_{max}=1e6,\rho=1.5, \varepsilon=1e-6, t=0.$\\
  \While{not converged \rm{or} $t>maxIter$}
 {
         Fix $\mathbf{b}_{k}^{t}, \mathbf{g}_{k}^{t}, \mathbf{h}_{k}^{t}$ to update $\mathbf{x}_{k}^{t+1}$ and $S_{k}^{t+1}$ by (22-24);\\
         Fix $\mathbf{x}_{k}^{t+1}, S_{k}^{t+1}, \mathbf{g}_{k}^{t}, \mathbf{h}_{k}^{t}$ to update $\mathbf{b}_{k}^{t+1}$ by (26);\\
         Fix $\mathbf{x}_{k}^{t+1}, S_{k}^{t+1}, \mathbf{b}_{k}^{t+1}, \mathbf{h}_{k}^{t}$ to update $\mathbf{g}_{k}^{t+1}$ by (28);\\
         Fix $\mathbf{x}_{k}^{t+1}, S_{k}^{t+1}, \mathbf{b}_{k}^{t+1}, \mathbf{g}_{k}^{t+1}$ to update $\mathbf{h}_{k}^{t+1}$ by (31);\\
         Update Lagrange multipliers by\
         $\pmb{\lambda}_{k}^{t+1}=\pmb{\lambda}_{k}^{t}+\mu^{t} (\mathbf{g}_{k}^{t+1}-\mathbf{x}_{k}^{t+1})$,\
         $\pmb{\nu}_{k}^{t+1}=\pmb{\nu}_{k}^{t}+\mu^{t} (\mathbf{h}_{k}^{t+1}-\mathbf{x}_{k}^{t+1})$;\\
         Update penalty parameter by
         $\mu^{t+1}=\min (\rho\mu^{t},\mu_{max})$;\\
         Check the convergence conditions:\\
         \eIf{$\norm {\mathbf{g}_{k}^{t+1}-\mathbf{x}_{k}^{t+1}}_{2}<\varepsilon$ and $\norm {\mathbf{h}_{k}^{t+1}-\mathbf{x}_{k}^{t+1}}_{2}<\varepsilon$ and $\norm {\mathbf{x}_{k}^{t+1}-\mathbf{x}_{k}^{t}}_{2}<\varepsilon$}
         {
           Stop iteration;
         }
         {
           $t\leftarrow t+1$;
         }
 }
 \end{spacing}
\end{algorithm}
\subsection{ADMM-based Optimization Algorithm}
In our case, the proposed ALMM framework can be roughly divided into two parts: {\it ALMM-based spectral unmixing (SU)} and {\it ALMM-based spectral variability dictionary learning (SVDL)}.
\subsubsection{ALMM-based Spectral Unmixing}
When $\mathbf{A}$ and $\mathbf{E}$ are given, Eq. (\ref{eq11}) is naturally converted into a problem of spectral unmixing as
\begin{equation}
\label{eq15}
\begin{aligned}
       \mathop{arg \min}_{\mathbf{X},\mathbf{S},\mathbf{B}} &\frac{1}{2}\norm{\mathbf{Y}-\mathbf{AXS}-\mathbf{EB}}_{\F}^{2}+\Phi(\mathbf{X})+\Psi(\mathbf{B})\\
      &\mathrm{s.t.} \quad \mathbf{X}\succeq 0,\quad \mathbf{S}\succeq 0.
\end{aligned}
\end{equation}
In order to conveniently and effectively collect all abundance vectors $\mathbf{X}$, we separately consider the problem (\ref{eq15}) over the $N$ pixels as pixel-wise spectral unmixing:
\begin{equation}
\label{eq16}
\begin{aligned}
       \mathop{arg \min}_{\mathbf{x}_{k},S_{k},\mathbf{b}_{k}} &\frac{1}{2} \norm{\mathbf{y}_{k}-(S_{k}\mathbf{A})\mathbf{x}_{k}-\mathbf{E}\mathbf{b}_{k}}_{2}^{2}+\Phi(\mathbf{x}_{k})+\Psi(\mathbf{b}_{k})\\
      &\mathrm{s.t.} \quad \mathbf{x}_{k}\succeq 0,\quad S_{k}\succeq 0.
\end{aligned}
\end{equation}
In \cite{Veganzones2014ELMM}, SCLSU is proposed to effectively solve the problem of scaled spectral unmixing. Equivalently, we formulate and solve the following nonnegative least squares (NNLS) problem \cite{Drumetz2016ELMM}\cite{Kim2010NNLS}:
\begin{equation}
\label{eq17}
\begin{aligned}
       \mathop{arg \min}_{\mathbf{x}_{k}\succeq 0} \frac{1}{2}\norm{\mathbf{y}_{k}-\mathbf{A}\mathbf{x}_{k}}_{2}^{2}.
\end{aligned}
\end{equation}
Once $\mathbf{x}_{k}$ is estimated by solving Eq. \eqref{eq17}, then $\mathbf{x}_{k}$ and $S_{k}$ can be simply derived, while satisfying the sum-to-one constraint with respect to $\mathbf{x}_{k}$ by
\begin{equation}
\label{eq18}
\begin{aligned}
       \hat{S}_{k}=\mathbf{1}^{T}\mathbf{x}_{k}, \quad  \hat{\mathbf{x}}_{k}=\mathbf{x}_{k}/\mathbf{1}^{T}\mathbf{x}_{k}.
\end{aligned}
\end{equation}
\begin{algorithm}[!t]
\footnotesize
\caption{ALMM-based SVDL}
\begin{spacing}{1}
\KwIn{$\mathbf{Y},\mathbf{A}$ and parameters $\alpha, \beta, \gamma, \eta, maxIter.$}
\KwOut{$\mathbf{E},\mathbf{X},\mathbf{S},\mathbf{B}.$}
\textbf{Initialization}: $ \mathbf{G}^{0}=\mathbf{H}^{0}=\mathbf{M}^{0}=\mathbf{0},\mathbf{S}^{0}=\mathbf{I}, \mathbf{B}^{0}=\mathbf{0},\pmb{\Delta}^{0}=\mathbf{0},$\
$\pmb{\Lambda}^{0}=\pmb{V}^{0}=\pmb{\Omega}^{0}=\mathbf{0},\mathbf{Q}^{0}=\mathbf{0}, \mathbf{T}^{0}=\mathbf{0}, \mathbf{\Pi}^{0}=\mathbf{0}, \mathbf{X}^{0}, \mathbf{E}^{0}, t=0,$\
$\xi^{0}=1e-3, \xi_{max}=1e6,\rho=1.5, \varepsilon=1e-6.$\\
\While{not converged \rm{or} $t>maxIter$}
 {
         Fix $\mathbf{E}^{t}, \mathbf{B}^{t}, \mathbf{X}^{t},\mathbf{S}^{t}$ to update $\mathbf{M}^{t+1}$ by (35);\\
         Fix $\mathbf{E}^{t}, \mathbf{M}^{t+1}$ to update $\mathbf{B}^{t+1}$  by (36);\\
         Fix $\mathbf{G}^{t}, \mathbf{H}^{t}, \mathbf{S}^{t}, \mathbf{M}^{t+1}, \pmb{\Lambda}^{t}, \pmb{V}^{t}$ to update $\mathbf{X}^{t+1}$ by (38-39);\\
         Fix $\mathbf{M}^{t+1}, \mathbf{X}^{t+1}, \mathbf{T}^{t}, \pmb{\Pi}^{t}, \pmb{\Delta}^{t}$ to update $\mathbf{S}^{t+1}$  by (41);\\
         Fix $\mathbf{M}^{t+1}, \mathbf{B}^{t+1}, \mathbf{Q}^{t}, \pmb{\Pi}^{t}$ to update $\mathbf{E}^{t+1}$  by (43);\\
         Fix $\mathbf{E}^{t}, \mathbf{E}^{t+1}, \pmb{\Pi}^{t}$ to update $\mathbf{Q}^{t+1}$  by (45);\\
         Fix $\mathbf{X}^{t+1}, \pmb{\Lambda}^{t}$ to update $\mathbf{G}^{(t+1)}$ by (46);\\
         Fix $\mathbf{X}^{t+1}, \pmb{V}^{t}$ to update $\mathbf{H}^{(t+1)}$ by (47);\\
         Fix $\mathbf{S}^{t+1}, \pmb{\Delta}^{t}$ to update $\mathbf{T}^{(t+1)}$ by (48);\\
         Update Lagrange multipliers by\\
              \hspace{0.8cm} $\pmb{\Lambda}^{t+1}\leftarrow\pmb{\Lambda}^{t}+\xi^{t} (\mathbf{G}^{t+1}-\mathbf{X}^{t+1})$\\
              \hspace{0.8cm} $\pmb{V}^{t+1}\leftarrow\pmb{V}^{t}+\xi^{t} (\mathbf{H}^{t+1}-\mathbf{X}^{t+1})$\\
              \hspace{0.8cm} $\pmb{\Omega}^{t+1}\leftarrow\pmb{\Omega}^{t}+\xi^{t} (\mathbf{M}^{t+1}-\mathbf{X}^{t+1}\mathbf{S}^{t+1})$\\
              \hspace{0.8cm} $\pmb{\Pi}^{t+1}\leftarrow\pmb{\Pi}^{t}+\xi^{t} (\mathbf{Q}^{t+1}-\mathbf{E}^{t+1})$\\
              \hspace{0.8cm} $\pmb{\Delta}^{t+1}\leftarrow\pmb{\Delta}^{t}+\xi^{t} (\mathbf{T}^{t+1}-\mathbf{S}^{t+1})$\\
         Update penalty parameter by
         $\xi^{(t+1)}=\min (\rho\xi^{(t)},\xi_{max})$;\\
         Check the convergence conditions:\\
         \eIf{$\norm {\mathbf{G}^{t+1}-\mathbf{X}^{t+1}}_{F}<\varepsilon$ and $\norm {\mathbf{H}^{t+1}-\mathbf{X}^{t+1}}_{F}<\varepsilon$ and $\norm {\mathbf{M}^{t+1}-\mathbf{X}^{t+1}\mathbf{S}^{t+1}}_{F}<\varepsilon$ and $\norm {\mathbf{Q}^{(t+1)}-\mathbf{E}^{(t+1)}}_{F}<\varepsilon$ and $\norm {\mathbf{T}^{t+1}-\mathbf{S}^{t+1}}_{F}<\varepsilon$ and $\norm {\mathbf{E}^{t+1}-\mathbf{E}^{t}}_{F}<\varepsilon$}
         {
           Stop iteration;
         }
         {
           $t\leftarrow t+1$;
         }
 }
 \end{spacing}
\end{algorithm}
In the following, we will effectively embed this idea into our framework to satisfy the sum-to-one constraint with respect to $\mathbf{x}_{k}$ and update $S_{k}$ simultaneously. Generally, the problem in (\ref{eq16}) can be seen as a Constrained Bilinear Regression Problem (CBRP). A similar CBRP has been effectively solved by the ADMM optimization algorithm \cite{Shekhar2013JointS}.

To facilitate an effective use of ADMM, we consider an equivalent form of (\ref{eq16}) by introducing multiple auxiliary variables $\mathbf{g}_{k}$, $\mathbf{h}_{k}$ to replace $\mathbf{x}_{k}$, $\mathbf{x}_{k}^{+}$, respectively, where $()^+$ denotes an operator that converts each component of the matrix to its absolute value, and $l_{R}^{+}(\mathbf{x})$ is defined as $\mathbf{x}\succeq \mathbf{0}$.
\begin{equation}
\label{eq19}
\begin{aligned}
      &\mathop{arg\min}_{\mathbf{x}_{k},S_{k},\mathbf{b}_{k},\mathbf{g}_{k},\mathbf{h}_{k}}\frac{1}{2}\norm{\mathbf{y}_{k}-(S_{k}\mathbf{A})\mathbf{x}_{k}-\mathbf{E}\mathbf{b}_{k}}_{2}^{2}\\
      & \qquad \qquad \quad +\Phi(\mathbf{g}_{k})+\Psi(\mathbf{b}_{k})+l_{R}^{+}(\mathbf{h}_{k})\\
      &\mathrm{s.t.} \quad S_{k}\succeq 0, \quad \mathbf{x}_{k}=\mathbf{g}_{k}, \quad \mathbf{x}_{k}^{+}=\mathbf{h}_{k},\quad \mathbf{h}_{k}\succeq \mathbf{0}.
\end{aligned}
\end{equation}
The augmented Lagrangian version of Eq. (\ref{eq19}) is
\begin{equation}
\label{eq20}
\begin{aligned}
      &\hspace{-0.4cm}\mathscr{L}_{U}\left(\mathbf{x}_{k},S_{k},\mathbf{b}_{k},\mathbf{g}_{k},\mathbf{h}_{k}, \pmb{\lambda}_{k}, \pmb{\nu}_{k} \right)\\
      &=\frac{1}{2}\norm{\mathbf{y}_{k}-(S_{k}\mathbf{A})\mathbf{x}_{k}-\mathbf{E}\mathbf{b}_{k}}_{2}^{2}+\Phi(\mathbf{g}_{k})+\Psi(\mathbf{b}_{k})\\
      &+\pmb{\lambda}_{k}^{T}(\mathbf{g}_{k}-\mathbf{x}_{k})+\pmb{\nu}_{k}^{T}(\mathbf{h}_{k}-\mathbf{x}_{k})+\frac{\mu}{2}\norm{\mathbf{g}_{k}-\mathbf{x}_{k}}_{2}^{2}\\
      &+\frac{\mu}{2}\norm{\mathbf{h}_{k}-\mathbf{x}_{k}}_{2}^{2}+l_{R}^{+}(\mathbf{h}_{k}),
\end{aligned}
\end{equation}
where $\pmb{\lambda}_{k}$, $\pmb{\nu}_{k}$, and $\pmb{\pi}_{k}$ are Lagrange multipliers and $\mu$ is the penalty parameter.The resulting algorithm of ALMM-based pixel-wise SU is detailed in \textbf{Algorithm 1}, and the solution to each subproblem is given in Appendix A. Correspondingly, variables $\mathbf{X},\mathbf{B}$, and $\mathbf{S}$ for all pixels can be collected using ALMM-based pixel-wise SU in turn.
\subsubsection{ALMM-based Spectral Variability Dictionary Learning}
If and only when $\mathbf{E}$ is unknown in Eq. (\ref{eq11}), we have to simultaneously perform spectral unmixing and dictionary learning using {\it ALMM-based SVDL}, resulting in alternately updating variables $\mathbf{X}$, $\mathbf{E}$, $\mathbf{B}$, and $\mathbf{S}$. This task essentially guides us to solve the optimization problem in Eq. (\ref{eq11}). Facing such a multi-variable optimization problem, we once again explore the ADMM algorithm for a fast and effective solution.It is noteworthy that concurrently estimating variables $\mathbf{X}$, $\mathbf{E}$, $\mathbf{B}$, and $\mathbf{S}$ in each iteration is of benefit to provide us a broader solution space and further find a better local minimum close to the global one easier.

By introducing multiple auxiliary variables $\mathbf{G}$, $\mathbf{H}$, $\mathbf{M}$, $\mathbf{T}$, and $\mathbf{Q}$ to replace $\mathbf{X}$, $\mathbf{X}^{+}$, $\mathbf{XS}$, $\mathbf{S}^{+}$, and $\mathbf{E}$, respectively, the augmented Lagrangian function of Eq. (\ref{eq11}) can be written as
\begin{equation}
\label{eq21}
\begin{aligned}
      &\hspace{-0.4cm}\mathscr{L}_{D} \left(\mathbf{X},\mathbf{S},\mathbf{E},\mathbf{B},\mathbf{G},\mathbf{H},\mathbf{M},\mathbf{T},\mathbf{Q},\pmb{\Lambda},\pmb{V},\pmb{\Omega}, \pmb{\Pi}, \pmb{\Delta} \right)\\
      &=\frac{1}{2}\norm{\mathbf{Y}-\mathbf{A}\mathbf{M}-\mathbf{E}\mathbf{B}}_{\F}^{2}+\Phi(\mathbf{G})+\Psi(\mathbf{B})+\Upsilon(\mathbf{Q})\\
      &+\pmb{\Lambda}^{\T}(\mathbf{G}-\mathbf{X})+\pmb{V}^{\T}(\mathbf{H}-\mathbf{X})+\pmb{\Pi}^{\T}(\mathbf{Q}-\mathbf{E})\\
      &+\pmb{\Omega}^{\T}(\mathbf{M}-\mathbf{XS})+\pmb{\Delta}^{\T}(\mathbf{\T}-\mathbf{S})+\frac{\xi}{2}\norm{\mathbf{G}-\mathbf{X}}_{\F}^{2}\\
      &+\frac{\xi}{2}\norm{\mathbf{H}-\mathbf{X}}_{\F}^{2}+\frac{\xi}{2}\norm{\mathbf{Q}-\mathbf{E}}_{\F}^{2}+\frac{\xi}{2}\norm{\mathbf{M}-\mathbf{XS}}_{\F}^{2}\\
      &+\frac{\xi}{2}\norm{\mathbf{T}-\mathbf{S}}_{\F}^{2}+l_{R}^{+}(\mathbf{H})+l_{R}^{+}(\mathbf{T}),\\
\end{aligned}
\end{equation}
where $\pmb{\Lambda}$, $\pmb{V}$, $\pmb{\Omega}$, $\pmb{\Pi}$, and $\pmb{\Delta}$ are Lagrange multipliers and $\xi$ is the penalty parameter.

The proposed algorithm for dictionary learning is summarized in \textbf{Algorithm 2} (see Appendix B for more details), where careful initialization is necessary in our case since the optimization problem of dictionary learning is not convex. The abundances generated by the SCLSU algorithm are set as the initial value ($\mathbf{X}^{0}$) and a random orthogonal matrix is produced to the initialization of the spectral variability dictionary ($\mathbf{E}^{0}$).
\begin{figure}[!t]
	  \centering
        \subfigure[ALMM-based SU]{
			\includegraphics[width=0.22\textwidth]{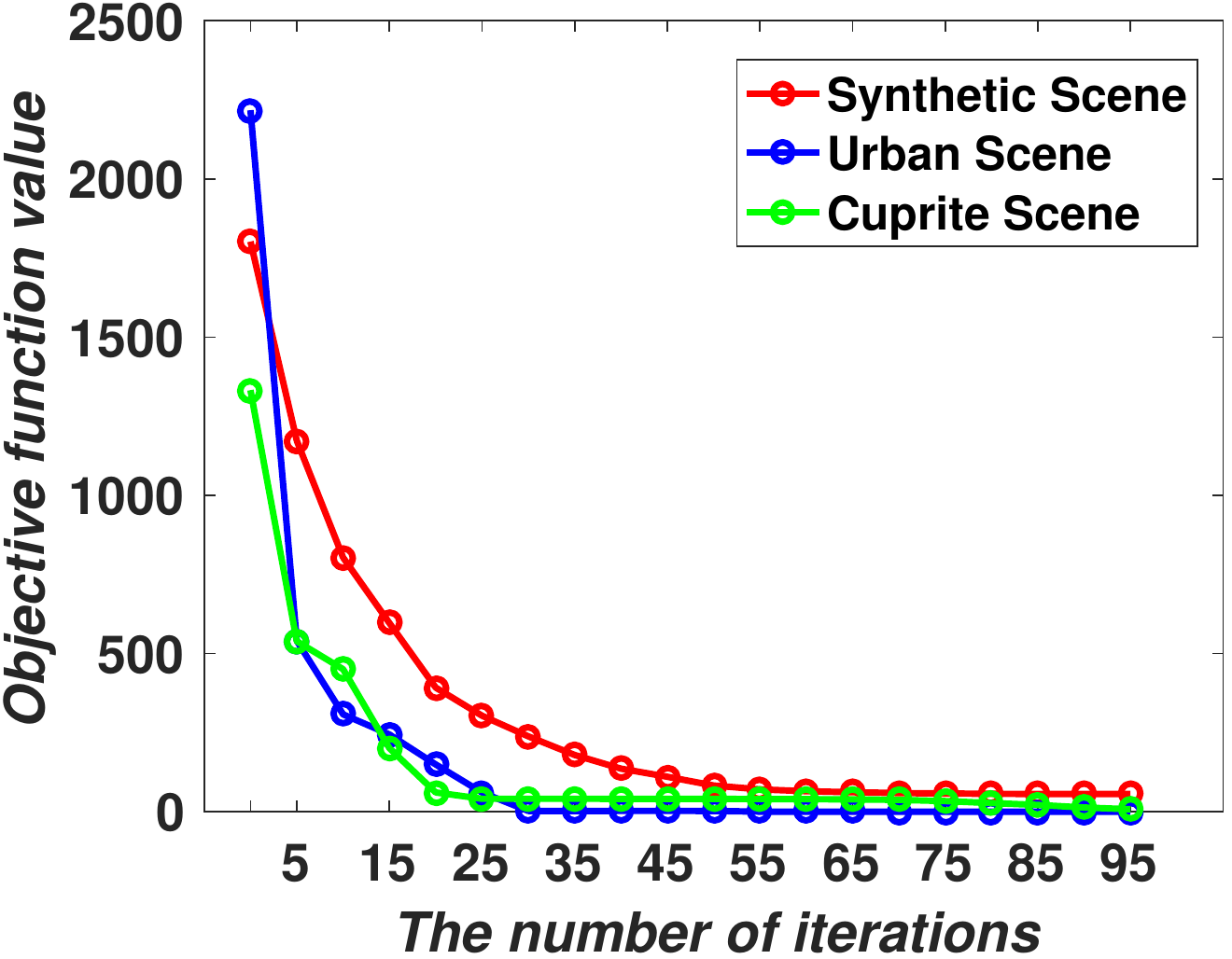}
            \label{fig:Convergence_ALMM_SU}
		}
		\subfigure[ALMM-based SVDL]{
			\includegraphics[width=0.22\textwidth]{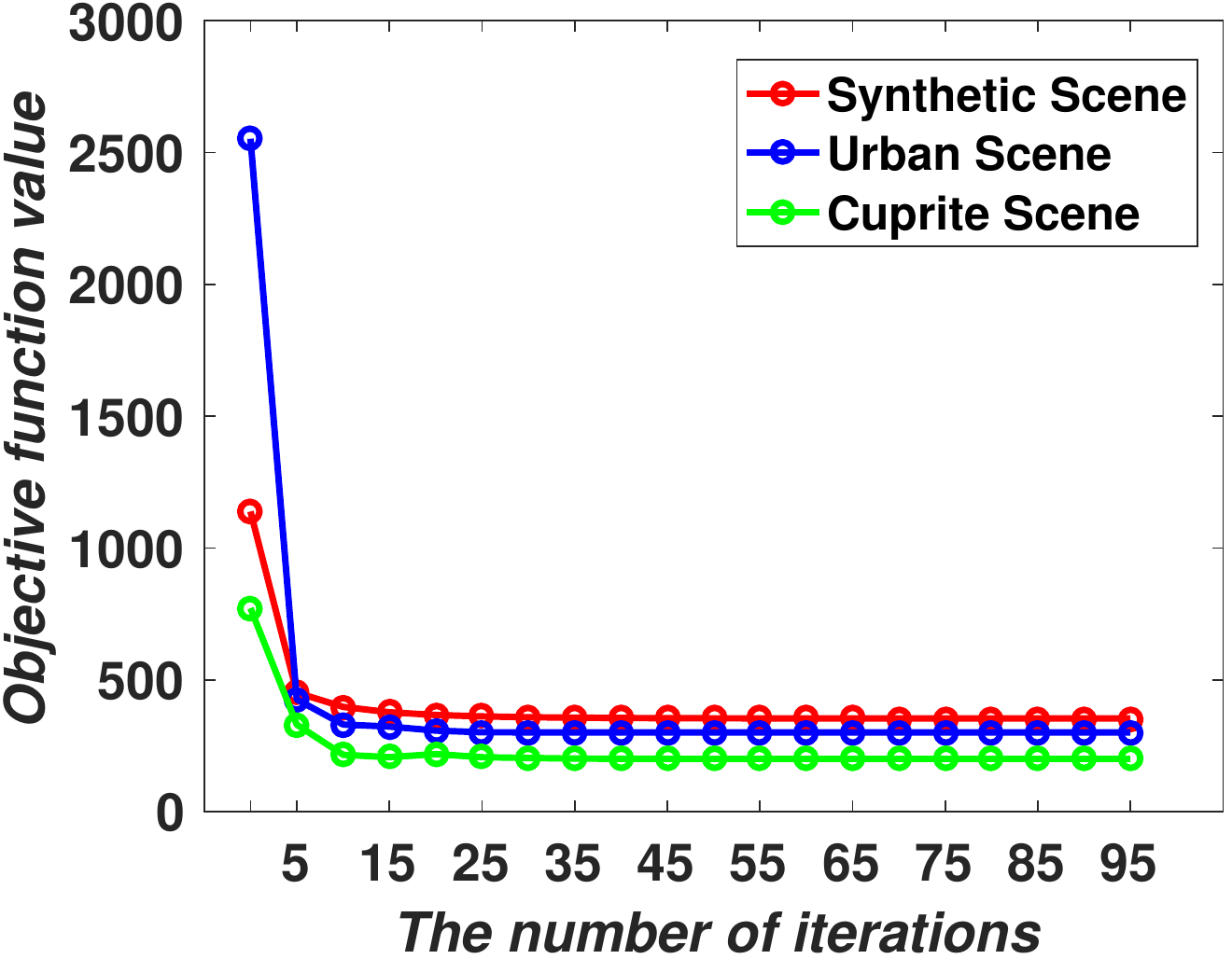}
            \label{fig:Convergence_ALMM_SVDL}
		}
         \caption{Convergence analysis of ALMM are experimentally performed on three different datasets.}
\label{fig:Convergence}
\end{figure}
\subsection{Convergence and Computational Cost}
The alternating scheme used in \textbf{Algorithm 1} and \textbf{Algorithm 2} is a typical multi-block ADMM optimization problem, whose convergence is theoretically supported in \cite{wang2015convergence}\cite{liu2017linearized}. Moreover, similar work for solving this sort of multi-block ADMM-based optimization problem has been successfully applied in \cite{Xu2012ADMM}\cite{Zhang2013MultiADMM}\cite{Zhou2017MultiADMM}. We experimentally visualize the convergence results for {\it ALMM-based SU} and {\it ALMM-based SVDL} on the three datasets, where the objective function value is recorded in each iteration (see Fig. \ref{fig:Convergence}). Notably, we collect the objective function values of all pixels computed by {\it ALMM-based pixel-wise SU} for obtaining {\it ALMM-based SU}'s.

We can clearly observe from Appendix A and B that the computational cost of our method is dominated by matrix products, yielding an overall $\mathcal{O}(DLN)$ w.r.t. {\it ALMM-based SU} and $\mathcal{O}(DL^{2}N)$ w.r.t. {\it ALMM-based SVDL}, respectively.
\section{Experiments}
\begin{figure}[!t]
	  \centering
			\includegraphics[width=0.45\textwidth]{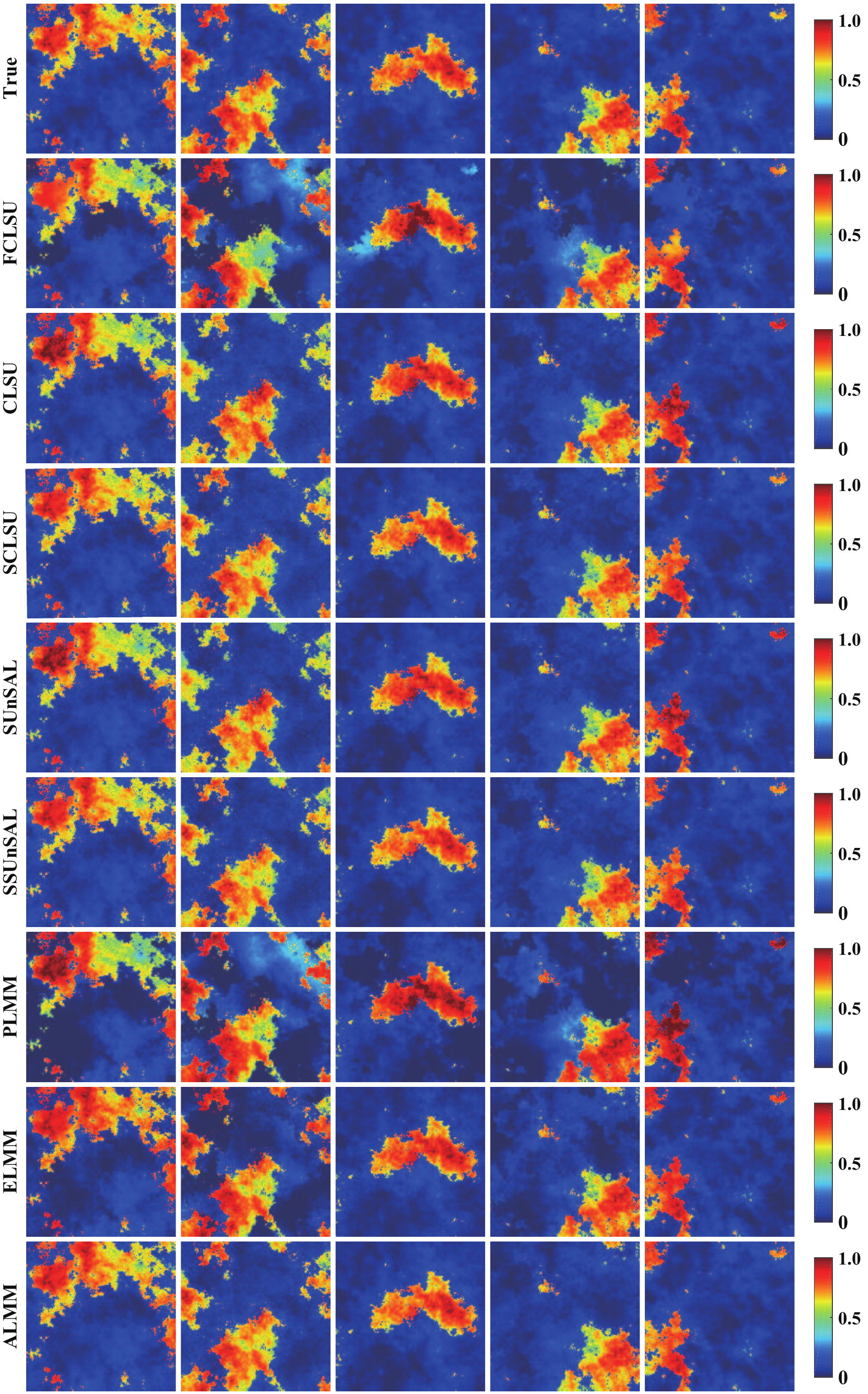}
         \caption{The abundances estimated by different SU methods (each column corresponds to one endmember extracted by VCA ) and the first row shows the ground truth.}
\label{fig:AbundancesSim}
\end{figure}

In this section, we quantitatively and visually evaluate the performance of the proposed method on three datasets: a synthetic dataset presented in \cite{Drumetz2016ELMM} and two real datasets over an urban area and the mining district in Cuprite, Nevada. We compare the proposed method (ALMM) with conventional and state-of-the-art approaches, including fully constrained least squares unmixing (FCLSU), constrained least squares unmixing (CLSU), scaled constrained least squares unmixing (SCLSU)\footnote{Without the term of $\mathbf{E}\mathbf{B}$, our model (ALMM) is equivalent to SCLSU.}, SUnSAL ($\ell_1$-CLSU), SSUnSAL (scaled SUnSAL), as well as PLMM and ELMM. Since the different regularization parameters lead to different results for each algorithm, we empirically and experimentally set up them to maximize performance. Specifically, we  set the penalty parameter of the sparsity-promoting term to be $6e-3$ in both SUnSAL and SSUnSAL, while three regularization parameters for abundances, endmembers, and perturbation in the PLMM are set to be $1e-2$, $1e-2$, and $1$, respectively. The regularization parameter $\lambda_{S}$ in the ELMM is set to be $0.5$. In the following experiments, we fix a display range for the abundance maps, e.g., [0, 1] for Fig. \ref{fig:AbundancesSim} and [0, 0.5] for Fig. \ref{fig:AbundanceCuprite}, in the interest of making fair visual comparisons. It should be noted that there are some abundances that show the maximum of the display range but actually exceed it, since they are generated by those algorithms without considering the scaling factors.
\begin{figure}[!t]
	  \centering
		\subfigure[A false color image]{
			\includegraphics[width=3cm,height=3cm]{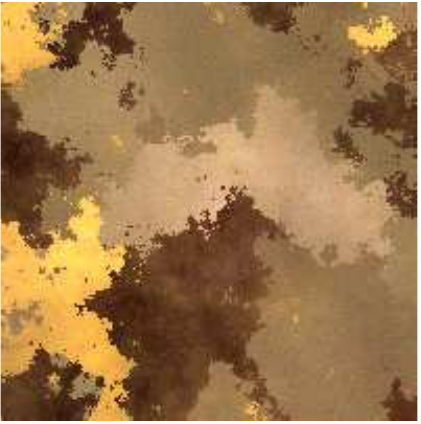}
		}
		\subfigure[Endmembers]{
			\includegraphics[width=4cm,height=3cm]{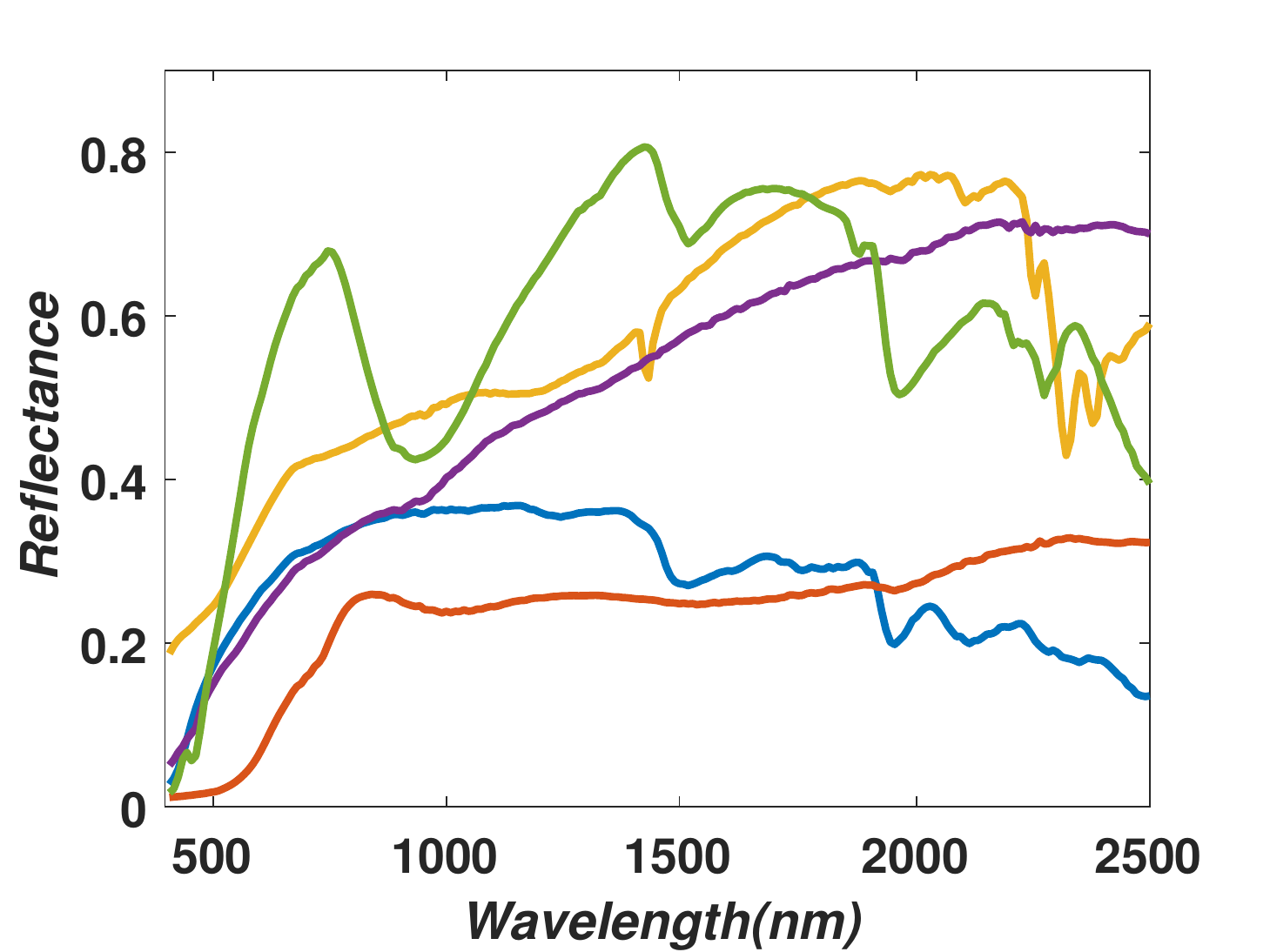}
		}

         \caption{A false color image of the synthetic data and five endmembers extracted by VCA.}
\label{fig:ImageSim}
\end{figure}
\begin{figure}[!t]
	  \centering
			\includegraphics[width=0.45\textwidth]{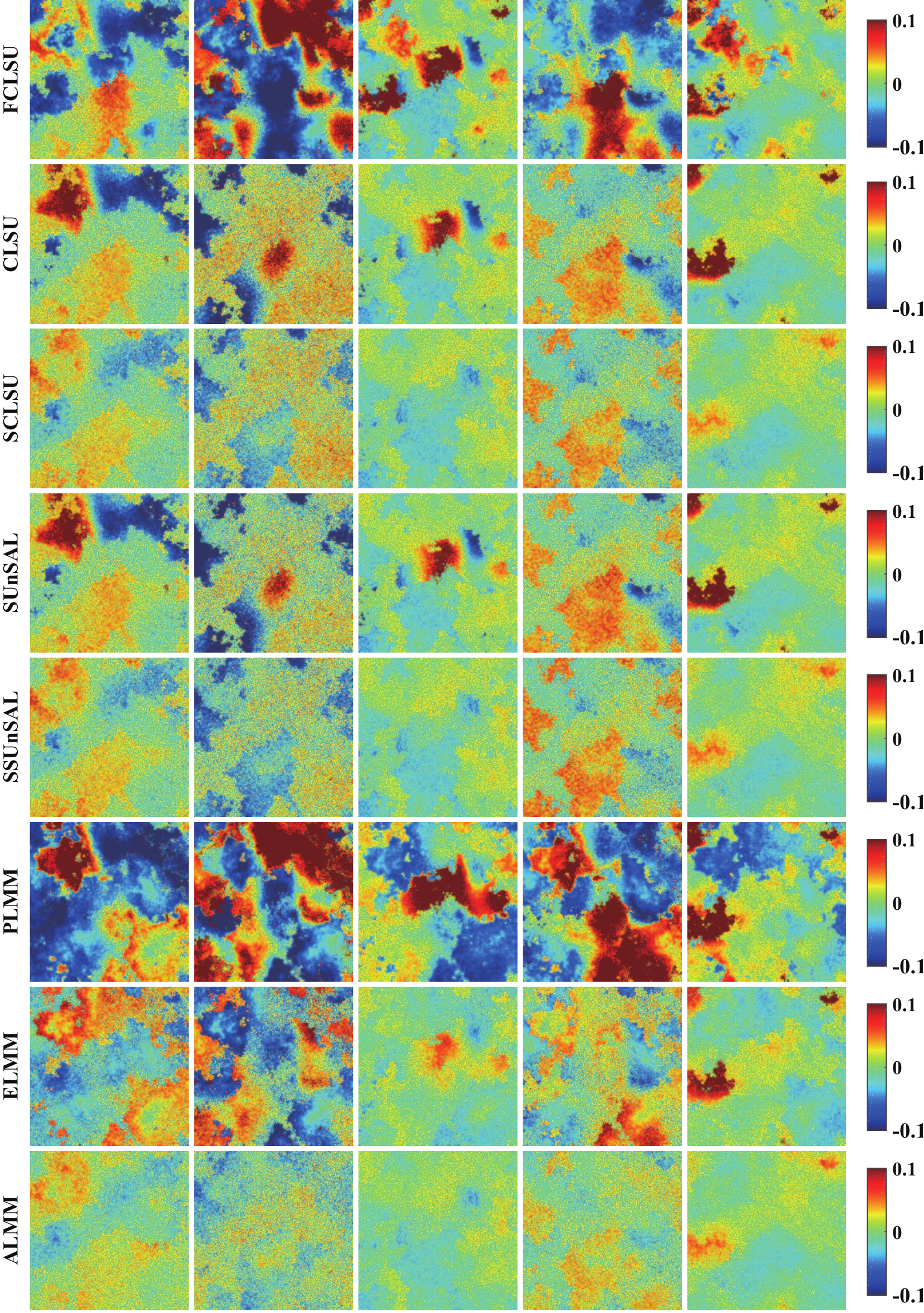}
         \caption{The difference abundance maps using different spectral unmixing methods corresponding to Fig. \ref{fig:AbundancesSim}.}
\label{fig:DiffAbundancesSim}
\end{figure}
\begin{figure*}[!t]
	  \centering
        \subfigure[Synthetic Scene]{
			\includegraphics[width=0.95\textwidth]{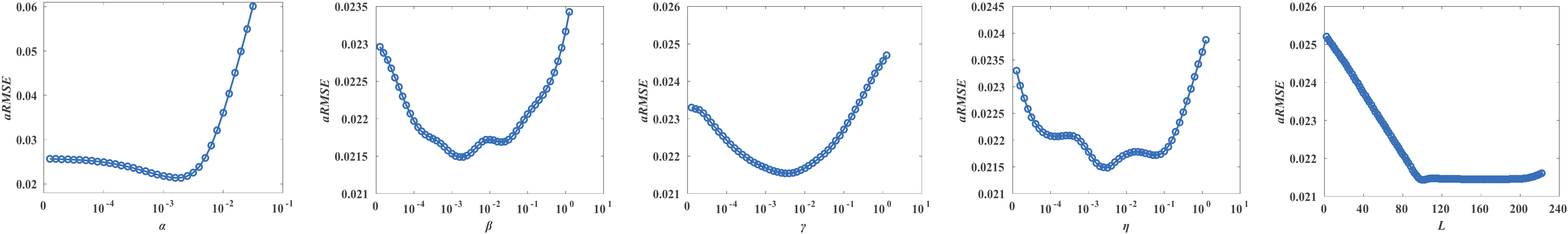}
		}
		\subfigure[Urban Scene]{
			\includegraphics[width=0.95\textwidth]{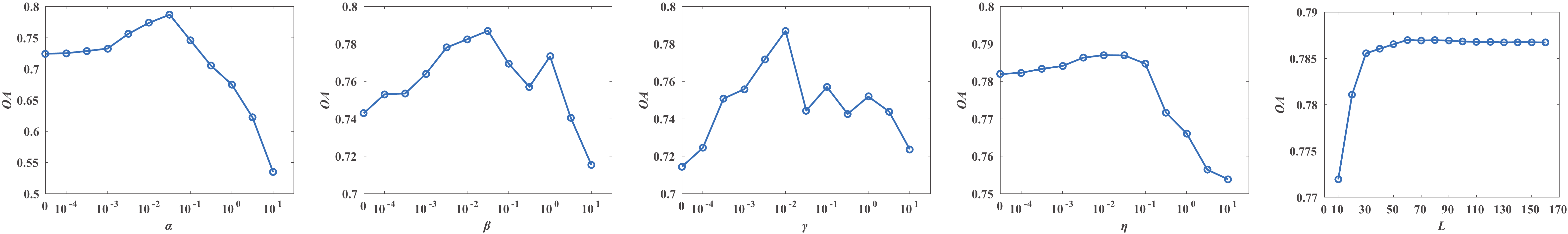}
		}
        \subfigure[Cuprite Scene]{
			\includegraphics[width=0.95\textwidth]{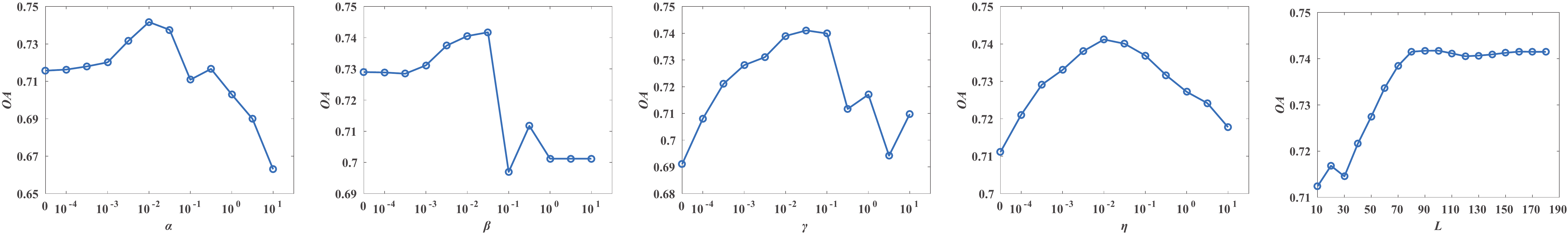}
		}
        \caption{Parameter sensitivity analysis of the proposed ALMM algorithm on three different study scenes for four regularization parameters: $\alpha$, $\beta$, $\gamma$, and $\eta$ as well as the number of basis vectors ($L$) of $\mathbf{E}$.}
\label{fig:Parameters}
\end{figure*}
\subsection{Synthetic Data}
\subsubsection{Data Description}
The synthetic data was simulated using five reference endmembers with 224 spectral bands randomly selected from the United States Geological Survey (USGS) spectral library and 200$\times$200 abundance maps generated using Gaussian fields, which were designed to satisfy the ANC and the ASC. Fig. \ref{fig:ImageSim} shows a false color image of the synthetic data and five selected endmembers. It should be noted that a spectral signature of each pixel in this dataset includes spectral variability due to endmember-dependent scaling factors and complex noise. Specifically, given five reference endmembers, we respectively multiply those spectral signatures by randomly-generated scaling factors ranging in [0.75, 1.25] and then a 25dB white Gaussian noise was added to these scaled reference endmembers. Next, we follow the LMM to mix them by means of generated abundance maps, finally a 25dB white Gaussian was added to these mixed pixels again. Following this simulation process, the generated spectral variabilities can be explained - without considering scaling factors - using a special mixtures of Gaussian distributions. Therefore, this simulated data with such spectral variability will give us a proper scenario to validate the proposed approach.
\subsubsection{Experimental Setup}
For a fair comparison, we adopt VCA to construct the endmember dictionary for all algorithms (including the proposed ALMM) under comparison, while Hysime \cite{BDias2008HySime} is used to estimate the number of endmembers. Endmember identification can be effectively performed with the spectral angle and five reference endmembers. Note that we show the averaged results for the different algorithms out of 10 runs, because VCA cannot always guarantee the same estimations in each round.

Importantly, a good initialization leads to a reasonable solution in our optimization problem due to nonconvexity. We hereafter initialize the abundance maps ($\mathbf{X}^{0}$) using the result of SCLSU. For the setting of other parameters, please refer to \textbf{Algorithm 1} and \textbf{Algorithm 2} for details.

For the performance assessment of the algorithms, we introduce three criteria to quantify experimental results: abundance overall root mean square error (aRMSE), reconstruction overall root mean square error (rRMSE), and average spectral angle mapper (aSAM). When the groundtruth of abundance maps is given, aRMSE can be defined as
\begin{equation}
\small
\label{eq22}
\begin{aligned}
       aRMSE=\frac{1}{N}\sum_{k=1}^N\sqrt{\frac{1}{P}\sum_{p=1}^P(\mathbf{x}_{kp}-\hat{\mathbf{x}}_{kp})^2}.
\end{aligned}
\end{equation}
Without the groundtruth of abundance maps, we can also give the two measures for assessing the performance of the algorithms from the point of view of data reconstruction. One of these measurements is rRMSE, defined by
\begin{equation}
\small
\label{eq23}
\begin{aligned}
       rRMSE=\frac{1}{N}\sum_{k=1}^N\sqrt{\frac{1}{D}\sum_{l=1}^D(\mathbf{y}_{kd}-\hat{\mathbf{y}}_{kd})^2},
\end{aligned}
\end{equation}
while the other is aSAM, expressed as
\begin{equation}
\small
\label{eq24}
\begin{aligned}
       aSAM=\frac{1}{N}\sum_{k=1}^Narccos\left(\frac{\mathbf{y}^{T}_{k}\hat{\mathbf{y}}_{k}}{\norm{\mathbf{y}_{k}}\norm{\hat{\mathbf{y}}_{k}}}\right).
\end{aligned}
\end{equation}
\textbf{Parameters Setting.} As the performance of the proposed ALMM model is fairly sensitive to the setting of four regularization parameters $\alpha$, $\beta$, $\gamma$, and $\eta$ as well as the number of basis vectors ($L$) of $\mathbf{E}$ , it is, therefore, indispensable to investigate the parameters setting in a proper range. For this reason, we attempt to find a group of stable and effective parameters by conducting several experiments on three different datasets (synthetic scene, urban scene, and Cuprite scene), as specifically shown in Fig. \ref{fig:Parameters} where we can empirically observe that $\alpha$ plays a dominant role in estimating the abundance maps ($\mathbf{X}$), while for other parameters ($\beta$, $\gamma$, $\eta$, and $L$) the importance of $L$ is visibly higher than that of the rest ones. With the increase of $L$, the algorithm performance is gradually improved until to around middle and then reaches a relatively stable state. An optimal performance can be obtained by setting these parameters as $\alpha=\beta=2e-3$, $\gamma=\eta=5e-3$, and $L=100$ in the synthetic scene, $\alpha=\beta=5e-2$, $\gamma=\eta=1e-2$, and $L=80$ in the urban scene, and $\alpha=1e-2$, $\beta=5e-2$, $\gamma=5e-2$, $\eta=1e-2$, and $L=90$ in the Cuprite scene. Accordingly, we can empirically summarize a general trend for the parameter-setting, that is, for regularization parameters ($\alpha$, $\beta$, $\gamma$, and $\eta$), they can be basically chosen in the range from $1e-3$ to $1e-2$, while $L$ tends to be approximately assigned to one half of the spectral length.
\begin{table*}[!t]
\begin{spacing}{1.3}
\centering
\caption{Quantitative performance comparison with the different algorithms on the synthetic data. The best one is shown in bold.}
\smallskip
\resizebox{\textwidth}{!}{
\begin{tabular}{|c|c|c|c|c|c|c|c|c|c|}
\hline Algorithm&FCLSU&CLSU&SUnSAL&SCLSU&SSUnSAL&PLMM&ELMM&ALMM\\
\hline aRMSE ($10^{-2}$)&6.30$\pm$0.70&4.21$\pm$0.20&3.99$\pm$0.33&2.63$\pm$0.23&2.43$\pm$0.16&6.21$\pm$0.43&3.23$\pm$0.28&\bf2.15$\pm$0.17\\
\hline rRMSE ($10^{-2}$)&1.50$\pm$0.024&1.23$\pm$0.0012&1.23$\pm$0.0012&1.23$\pm$0.0012&1.23$\pm$0.0012&1.29$\pm$0.0037&0.58$\pm$0.0005&\bf0.018$\pm$0.00004\\
\hline aSAM ($10^{-2}$)&198.36$\pm$2.54&177.17$\pm$0.016&177.26$\pm$0.018&177.17$\pm$0.016&177.26$\pm$0.018&184.27$\pm$1.09&83.92$\pm$0.007&\bf1.04$\pm$0.0002\\
\hline
\end{tabular}
}
\label{tab:EvaluationSim}
\end{spacing}
\end{table*}
\begin{figure*}[!t]
	  \centering
		\subfigure{
			\includegraphics[width=16cm,height=8cm]{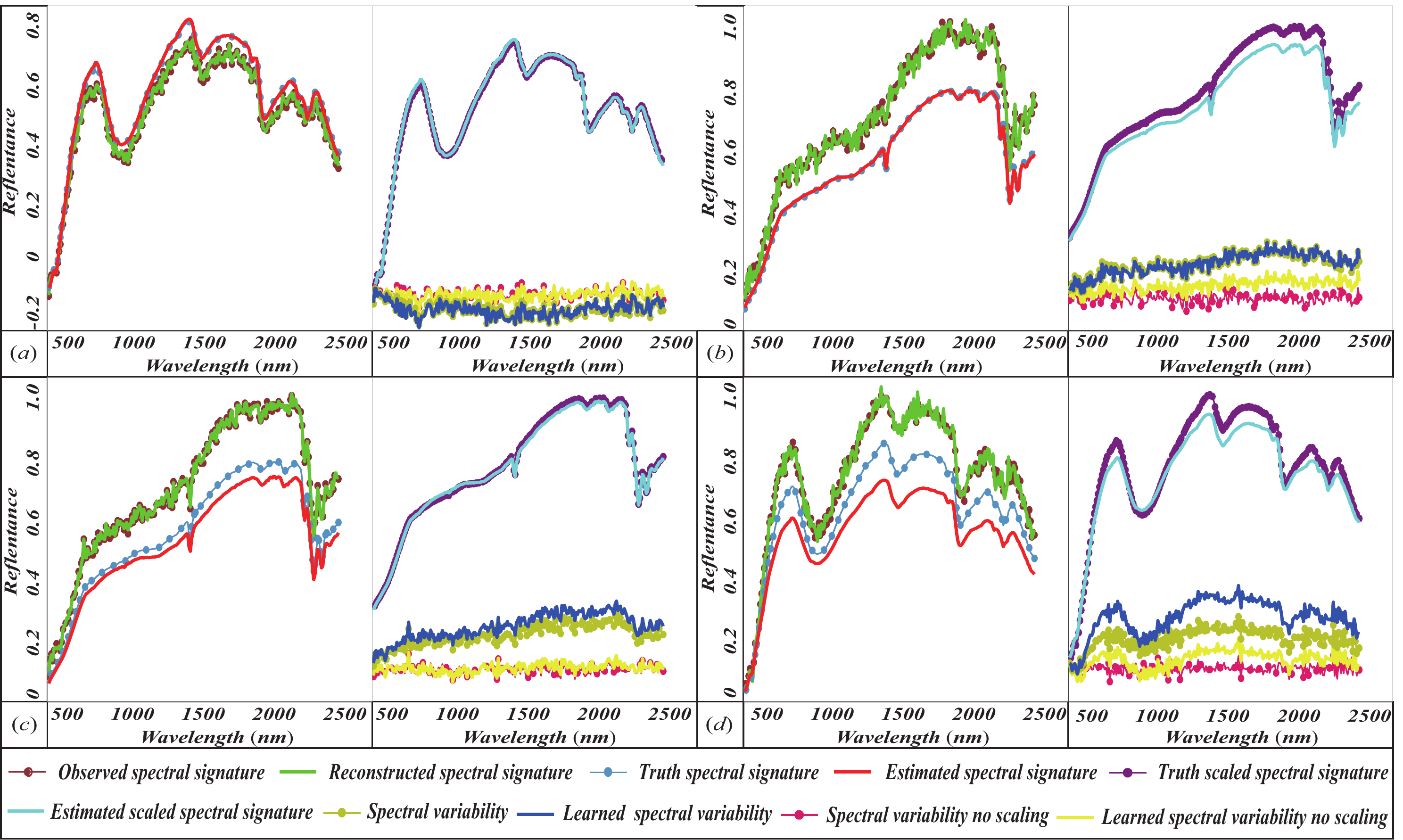}
		}
         \caption{Highlight some typical cases with respect to learned spectral variabilities. (a): ALMM not only can reconstruct the observed spectral signature well but also learn the various spectral variabilities (e.g., scaling factors and other complex variabilities) effectively. (b): In this case, although the scaling factors fail to be estimated well, yet the term of $\mathbf{E}\mathbf{B}$ effectively fix the errors, still leading to a desirable abundances estimation. (c) shows a bad example in estimating scaling factors, while (d) gives a failure case of learning spectral variabilities. Please refer to the fifth paragraph in Section IV.A(3) for more analysis and discussion.}
\label{fig:exampleLSV}
\end{figure*}
\subsubsection{Results and Analysis}
Fig. \ref{fig:AbundancesSim} shows the estimated abundance maps for the aforementioned algorithms, while Table \ref{tab:EvaluationSim} details the corresponding quantitative assessment results. Since the visual difference of the estimated abundance maps is not obvious among some of the algorithms, the abundance difference maps are also given in Fig. \ref{fig:DiffAbundancesSim} to intuitively highlight the difference.

As can be seen in Table I, FCLSU yields a poor performance due to the presence of spectral variability. CLSU performs better than FCLSU since the abundances can be reasonably estimated in a cone not in a simplex by dropping the ASC. However, the spectral variability is not actually eliminated by CLSU, but rather absorbed by the abundances. Fig. \ref{fig:AbundancesSim} provides good evidence that the abundance values for some pixels are higher than 1, which violates the ASC. With the consideration of the ASC, SCLSU performs better than CLSU, particularly being robust against scaling factors. By adding the sparsity term, SUnSAL and SSUnSAL can further improve the performance compared to CLSU and SCLSU without sparsity term, which experimentally explains that each pixel in the studied hyperspectral scene consists of a few materials.

Although the ELMM approach can model scaling variability with reasonable physical consideration, the difficult parameter estimation results in its limited performance. More specifically, the objective function of the ELMM is an obvious non-convex problem, since the scaling factors and abundance maps need to be estimated simultaneously, which easily drops to a local minimum and leads to the  inaccurate estimation of the abundances and scaling factors. Generally, the scaling factors among different endmembers are highly correlated in reality, because the endmember variability is often dominated by the geometry effect; this is another factor that hinders the improvement of the performance of the ELMM. PLMM fails to specify the spectral variabilities (e.g., scaling factors) according to their properties.

By comparison, the proposed method outperforms other algorithms, which suggests that this method can effectively learn the spectral variability, improving the accuracy of the abundance estimation. Fig. \ref{fig:DiffAbundancesSim} illustrates a more significant comparison by means of abundance difference maps between the groundtruth and estimated abundance maps of the compared algorithms. The difference values obtained from ALMM are mostly close to zero, which indicates that the performance of ALMM is superior to that of the other methods.

For the purpose of highlighting the learnability for spectral variabilities, we emphatically investigate several typical examples of learned spectral variabilities by giving the four spectral signatures under different conditions, as shown in Fig. \ref{fig:exampleLSV}, including
\begin{itemize}
\item the observed spectral signature ($\mathbf{y}_{k}$) and the reconstructed spectral signature ($\mathbf{A}\hat{\mathbf{x}}_{k}\hat{S}_{k}+\hat{\mathbf{E}}\hat{\mathbf{b}}_{k}$),
\item the truth spectral signature ($\mathbf{A}\mathbf{x}_{k}$) and the estimated spectral signature ($\mathbf{A}\hat{\mathbf{x}}_{k}$),
\item the truth scaled spectral signature ($\mathbf{A}\mathbf{x}_{k}S_{k}$) and the estimated scaled spectral signature ($\mathbf{A}\hat{\mathbf{x}}_{k}\hat{S}_{k}$),
\item spectral variability ($|1-S_{k}|\mathbf{A}\mathbf{x}_{k}+\mathbf{E}\mathbf{b}_{k}$) and learned spectral variability ($|1-\hat{S}_{k}| \hat{\mathbf{A}}\hat{\mathbf{x}}_{k}+\hat{\mathbf{E}}\hat{\mathbf{b}}_{k}$) and
\item spectral variability without scaling ($\mathbf{E}\mathbf{b}_{k}$) and learned spectral variability without scaling ($\hat{\mathbf{E}}\hat{\mathbf{b}}_{k}$).
\end{itemize}
Using ALMM, scaling factors can be approximately fit by $S_{k}$ , as shown in Fig. \ref{fig:exampleLSV}(a), while for those spectral variabilities that cannot be fully explained by scaling factors, $\mathbf{E}\mathbf{b}_{k}$ can correct them (see Fig. \ref{fig:exampleLSV}(b)). More discussion can be detailed as follows
\begin{itemize}
\item Fig. \ref{fig:exampleLSV}(a) shows the expected competitive result: the spectral signature and learned spectral variabilities basically match the real ones. We have to point out that most of the pixels in this simulated data follow the expected results.
\item Fig. \ref{fig:exampleLSV}(b) shows another case where each endmember in the given mixed pixel is not sharing similar scalar, so it would fail to estimate the scaling factors accurately. In such a case, however, the abundance map can be estimated well, since the spectral variability term ($\mathbf{E}\mathbf{b}_{k}$) can effectively represent the rest of the spectral variabilities that cannot be explained by a shared scaling factor, as displayed in the curves of the second figure of Fig. \ref{fig:exampleLSV}(b).
\item As can be seen in Fig. \ref{fig:exampleLSV}(c), although our model can learn spectral variability without scaling factors well, it fails to effectively estimate the truth spectral signature, further causing inaccurate abundance maps. This probably results from the inaccurate estimation of scaling factors.
\item A counterexample that cannot handle the spectral variability is given in Fig. \ref{fig:exampleLSV}(d). Such a negative example is unexpected but reasonable due to non-convexity, which masks it difficult for our model to precisely estimate all variables. Although proper prior assumptions and the endmember dictionary extracted by VCA are used in our model, they can only shrink the range of solutions rather than giving the globally optimal solution directly.
\end{itemize}
\begin{table*}[!t]
\begin{spacing}{1.3}
\centering
\caption{Quantitative performance comparison with the different algorithms on the Urban data. The best one is shown in bold.}
\smallskip
\resizebox{\textwidth}{!}{
\begin{tabular}{|c|c|c|c|c|c|c|c|c|c|}
\hline Algorithm&FCLSU&CLSU&SUnSAL&SCLSU&SSUnSAL&PLMM&ELMM&ALMM\\
\hline OA (\%)&54.66$\pm$7.59&68.08$\pm$5.24&71.26$\pm$5.50&68.08$\pm$5.24&71.26$\pm$5.50&58.55$\pm$7.21&62.41$\pm$6.87&\bf78.70$\pm$2.83\\
\hline rRMSE ($10^{-2}$)&3.97$\pm$0.32&0.86$\pm$0.085&0.86$\pm$0.085&0.86$\pm$0.085&0.86$\pm$0.085&1.11$\pm$0.20&0.72$\pm$0.048&\bf0.27$\pm$0.0003\\
\hline aSAM ($10^{-2}$)&867.34$\pm$50.14&295.69$\pm$26.19&295.82$\pm$32.03&295.69$\pm$26.19&295.82$\pm$32.03&362.40$\pm$60.04&203.78$\pm$19.95&\bf58.02$\pm$1.81\\
\hline
\end{tabular}}
\label{tab:EvaluationUrban}
\end{spacing}
\end{table*}
\begin{figure}[!t]
	  \centering
		\subfigure[Single Gaussian noise (different SNRs)]{
			\includegraphics[width=4.1cm,height=3cm]{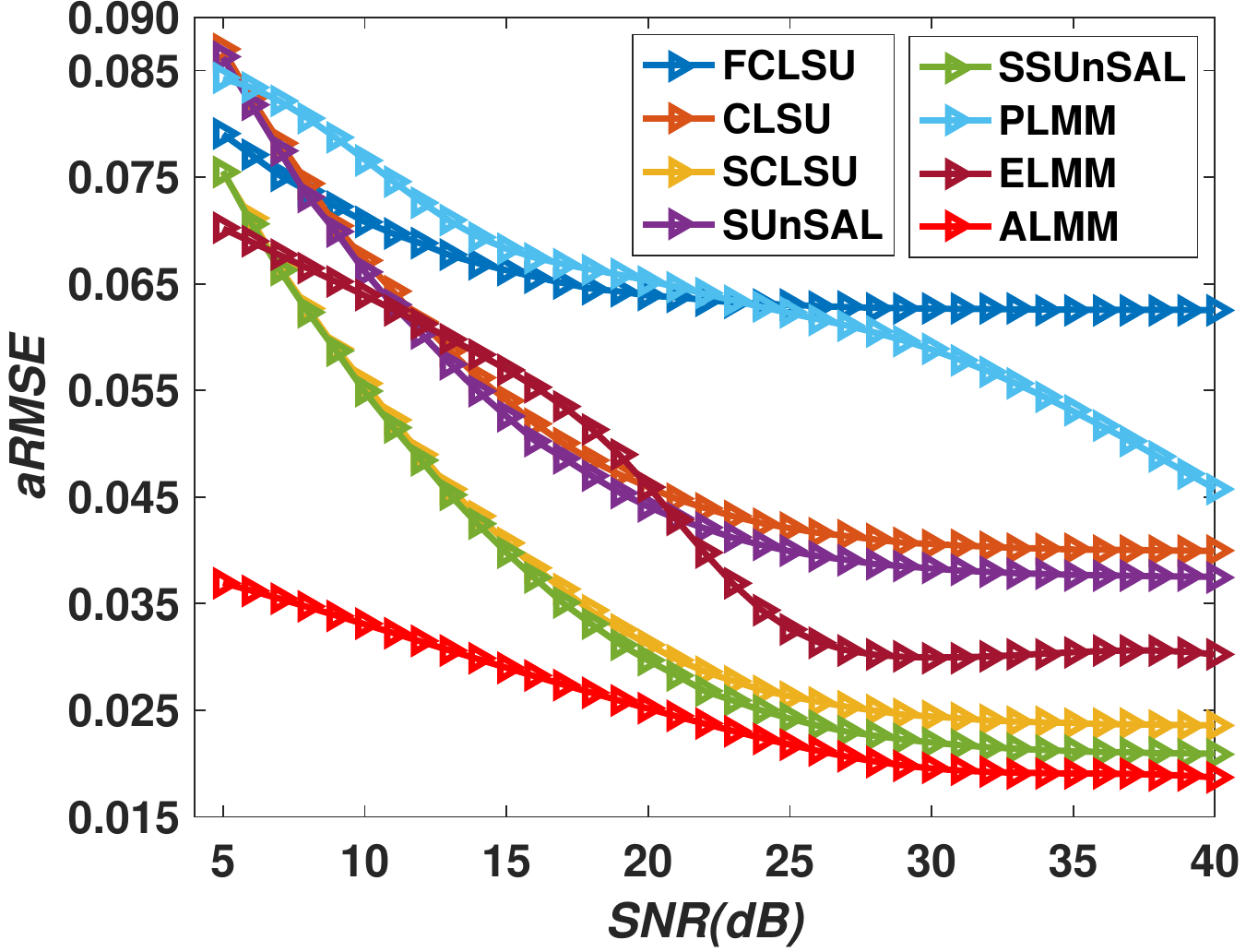}
            \label{fig:Robustnessa}
		}
		\subfigure[Mixtures of Multiple Gaussian noises]{
			\includegraphics[width=4.1cm,height=3cm]{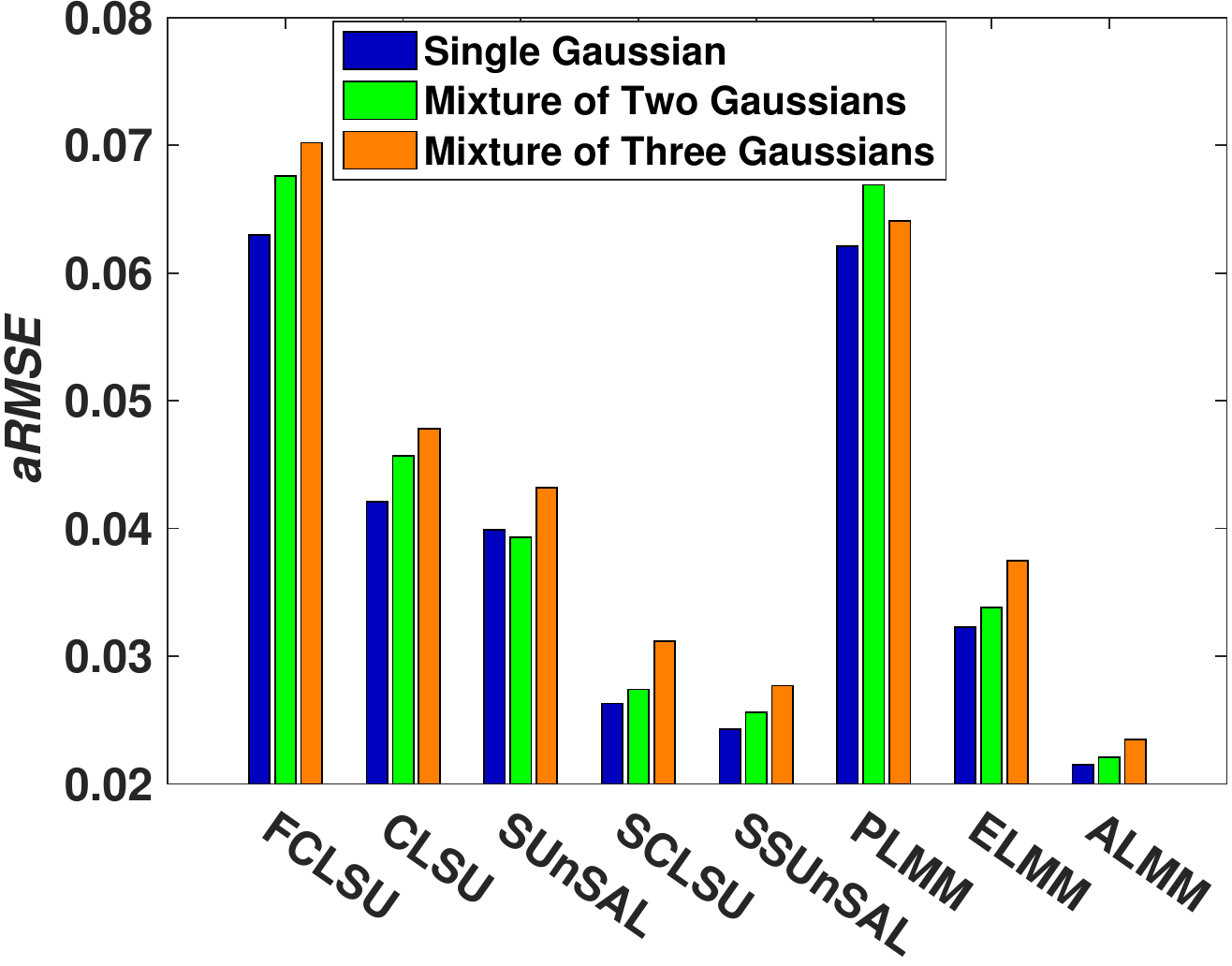}
            \label{fig:Robustnessb}
		}
         \caption{Robustness analysis using single Gaussian noise with the different SNRs and mixtures of multiple Gaussian noises.}
\label{fig:Robustness}
\end{figure}
\subsubsection{Robustness Study}
To quantitatively validate the robustness of our method, we investigate the performances (aRMSEs) of the different algorithms on simulated data by adding Gaussian white noises with the different signal-to-noise-ratio (SNR) ranging from 5dB to 40dB at a 1dB interval. As can be clearly seen from Fig. \ref{fig:Robustnessa} that ALMM is more robust and effective against noises with the different SNRs, compared to others. Also, we experimentally discuss another case of mixed Gaussian distributions as the noise input. More specifically, mixtures of Gaussian distributions can be generated by assembling several single Gaussian distributions with the different mean and variance randomly selected from $0$ to $0.01$. Using them (single Gaussian, mixtures of two Gaussian, and mixtures of three Gaussian), we horizontally compare the performance of different algorithms by the averaged aRMSEs out of 20 runs to achieve the reliable results. It is clear in Fig. \ref{fig:Robustnessb} that the ALMM performs better and more robust against the Gaussian mixture noises than other comparative algorithms.
\subsection{First Real Data (Urban)}
\subsubsection{Data Description}
This dataset was collected by the Hyperspectral Digital Imagery Collection Experiment (HYDICE) over an urban area at Copperas Cove, Texas, USA. The dataset has been widely used in the field of hyperspectral unmixing \cite{Zhu2014Urban}\cite{Wang2015UrbanTIP}\cite{Liu2011SU}. The latest data version was issued by Geospatial Research Laboratory (USA) and Engineer Research and Development Center (USA) in $2015$.\footnote{http://www.tec.army.mil/Hypercube} The image consists of $307 \times 307$ pixels with $210$ spectral bands in the wavelength from 400 nm to 2500 nm with spectral resolution of 10 nm at a ground sampling distance (GSD) of 2 m. Fig. \ref{fig:ImageUrbana} shows a false color image of the study scene. Due to water absorption and atmospheric effects, we reduced $210$ bands to $162$ bands by removing bands $1$-$4$, $76$, $87$, $101$-$111$, $136$-$153$, and $198$-$210$.
\subsubsection{Experimental Setup}
\begin{figure}[!t]
	  \centering
		\subfigure[A false color image]{
			\includegraphics[width=3cm,height=3cm]{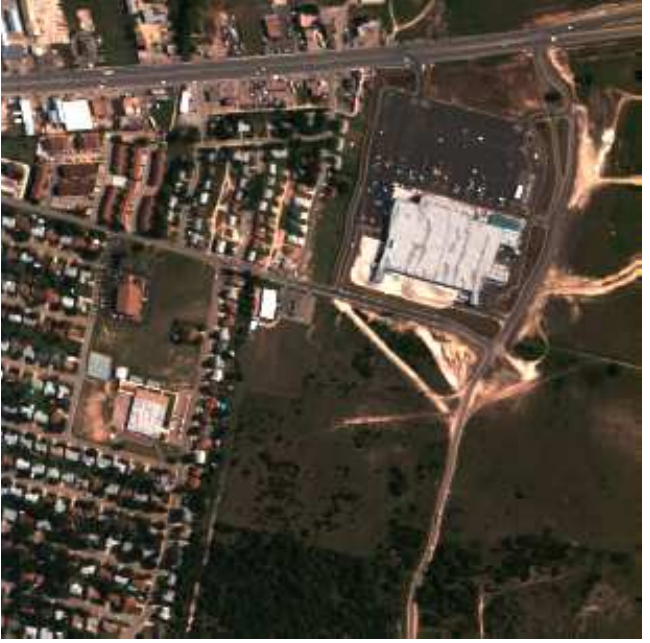}
            \label{fig:ImageUrbana}
		}
		\subfigure[Endmembers]{
			\includegraphics[width=4cm,height=3cm]{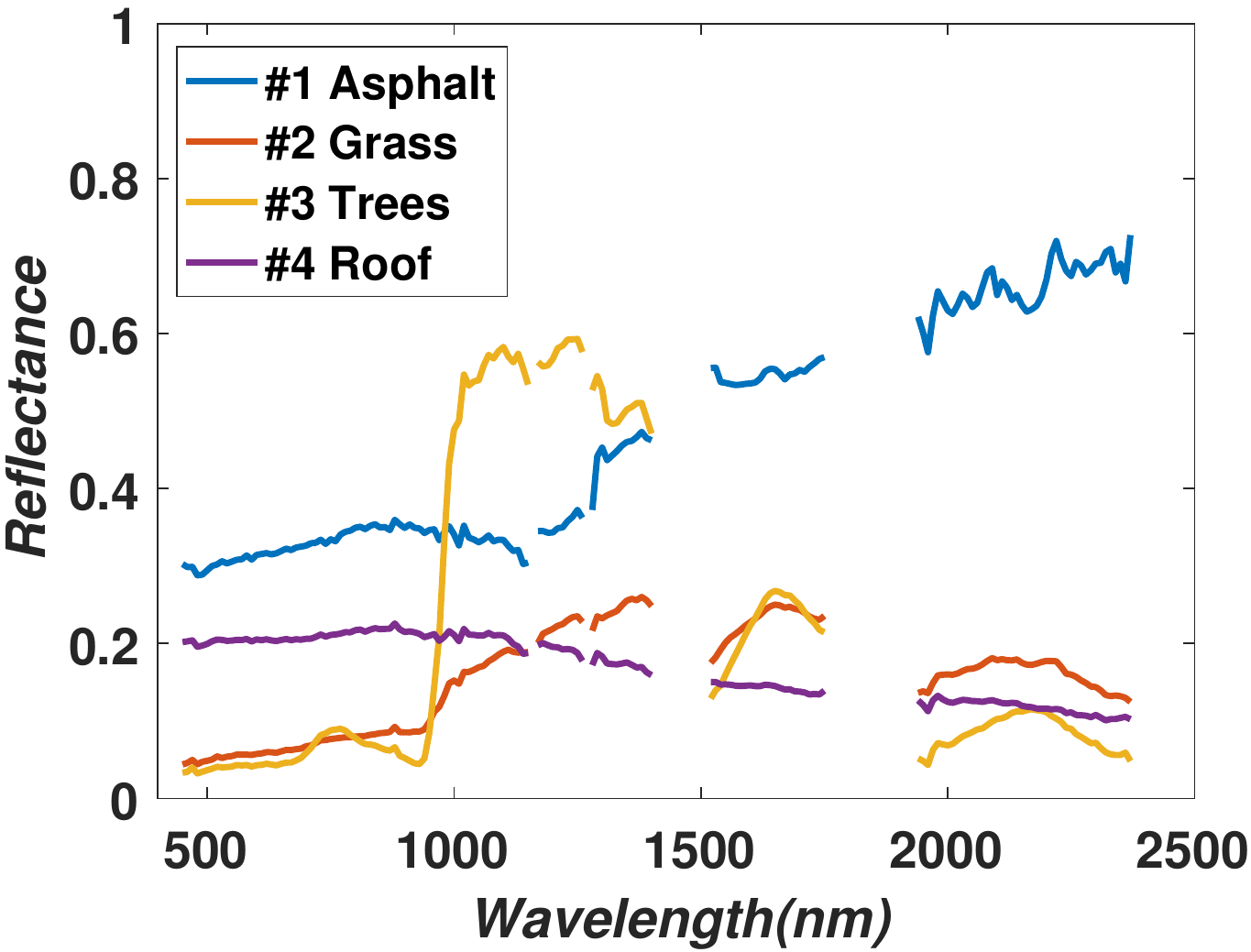}
            \label{fig:ImageUrbanb}
		}
         \caption{A false color image of the Urban data and four endmembers used in spectral unmixing.}
\label{fig:ImageUrban}
\end{figure}

Four main endmembers can be observed in the scene: asphalt (road and parking lot), grass, trees, and roof. For more discussion and analysis regarding these endmembers, refer to \cite{Zhu2014Urban}\cite{Liu2011SU}. Likewise, VCA and HySime are adopted to build the endmember dictionary and determine the number of endmembers for all algorithms (including ALMM), respectively. Fig. \ref{fig:ImageUrbanb} shows the endmembers used in spectral unmixing. Furthermore, the material identification step is performed through comparison with the reference endmembers. \footnote{The reference endmembers are manually extracted from the original image. Please refer to \cite{Liu2011SU} and \cite{Wang2015UrbanTIP} for details.}
\subsubsection{Results and Analysis}
\begin{figure}[!t]
	  \centering
			\includegraphics[width=0.45\textwidth]{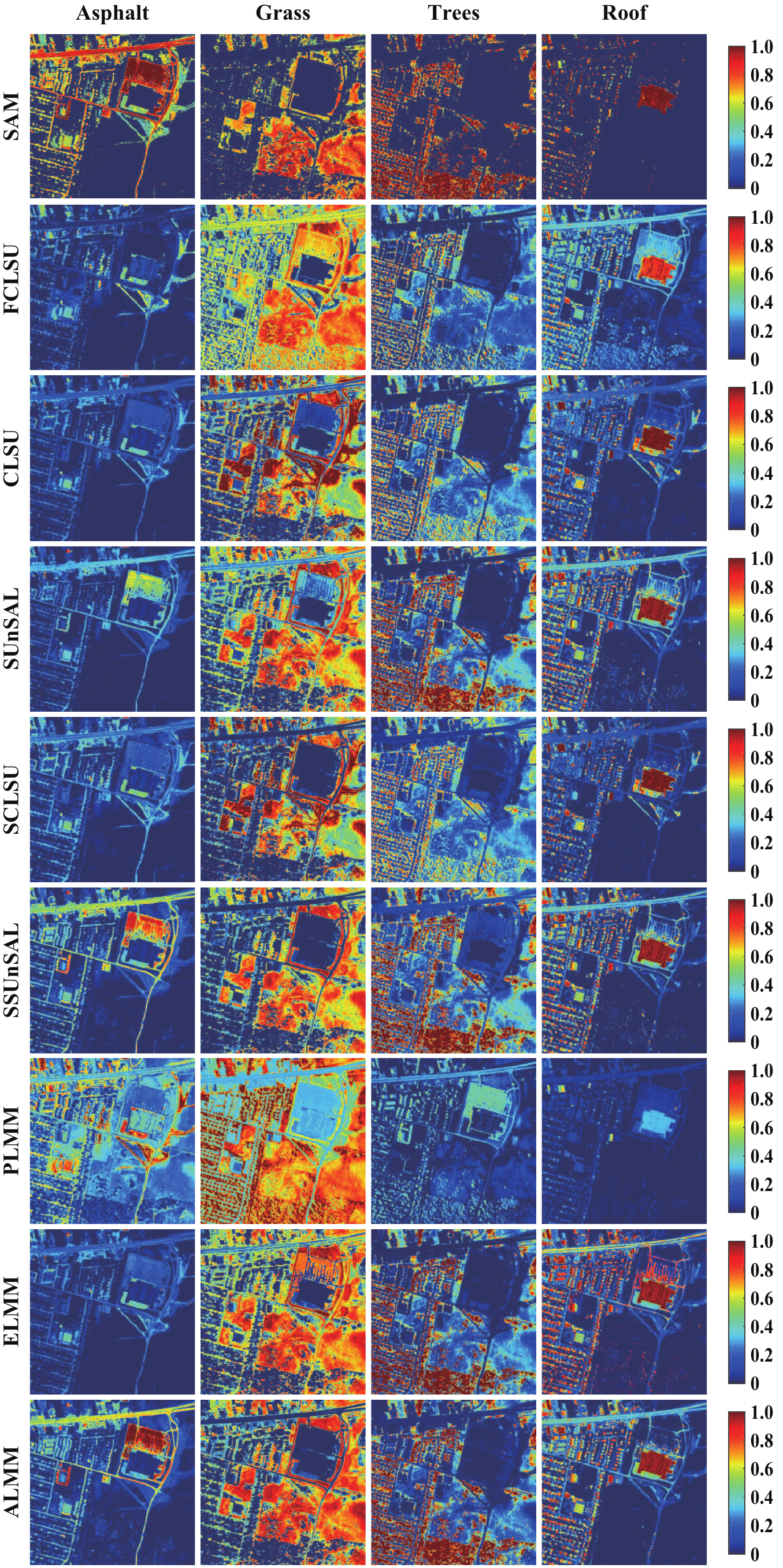}
         \caption{The abundance maps comparison between the proposed method and the state-of-art methods.}
\label{fig:AbundancesUrban}
\end{figure}

For the quantitative assessment of the experimental results, we calculate the two indices, rRMSE and aSAM. Since there is no groundtruth of the abundance maps for the real data and meanwhile the metrics based on reconstruction errors (rRMSE and aSAM) are not suitable to assess the performance of spectral unmixing. For these reasons, we propose a classification-based evaluation strategy for assessing the abundance maps using the overall accuracy (OA). Firstly, we perform the spectral angle mapper (SAM) classification using the reference endmembers as reference spectra. The first row of Fig. \ref{fig:AbundancesUrban} shows the cosine similarity for the four classes, where negative samples are masked out with 0. For the spectral unmixing results, we obtain classification maps by classifying each pixel into an endmember that has the maximum abundance value. By using the SAM classification result as the groundtruth, OA can be calculated for the different methods, as listed in Table II.

We also perform a visual examination to evaluate the performance of the algorithms for the estimation of abundance maps. According to the quantitative and visual results, we analyze the performance of the different algorithms as follows. FCLSU performs rather poor estimation for the abundances, since spectral variability comes into play in the real data. Similarly, CLSU also fails to deal with spectral variability; however it outperforms FCLSU, as shown visually in the Fig.  \ref{fig:AbundancesUrban} and Table \ref{tab:EvaluationUrban}, due to the relaxation of the ASC. When scaling factors are considered, there is better identification of the materials of asphalt, trees, and roof using SCLSU. In particular, FCLSU and CLSU both fail to detect the material of asphalt, but SCLSU effectively does.

Although ELMM is able to detect some areas where only one scaling factor presents a difficulty for interpreting all endmembers and meanwhile obtains a relatively lower rRMSE and aSAM as listed in Table II, the non-convexity involved in the simultaneous estimation of the abundance maps and scaling factors prevents ELMM from achieving better performance (lower CMMS than that of CLSU and SCLSU). As shown in the third row of Fig. \ref{fig:AbundancesUrban}, ELMM obtains a purer identification for the materials of trees and roof, while there is still room for improvement in its abundance estimation of asphalt and grass. In Fig.  \ref{fig:AbundancesUrban}, the performance of PLMM is relatively poor because it is not able to address the scaling factors, which is the main spectral variability in the study scene.

In this scene, there are many pure pixels, owing to high resolution; however, they are considered mixed pixels in the comparison of methods due to the existence of spectral variability. As shown in Fig. \ref{fig:AbundancesUrban}, the visual performance of the proposed ALMM method is superior to the other methods and consistent numerical evaluation is listed in Table II as well. More specifically, the asphalt is purely identified by ALMM, unlike the others; and a similar observation can be found in the grass as well. For the trees and the roof, the abundance maps estimated by ALMM show higher contrast than those estimated by other methods. This result implies that the proposed method successfully addresses spectral variability.

In order to further highlight the differences between the proposed method and CLSU, SCLSU, SUnSAL, and SSUnSAL, we emphatically focus on their abundance maps. Each material, by and large, becomes more purely identified and the corresponding abundance maps clearer by successively using CLSU, SCLSU, SUnSAL, SSUnSAL, and the proposed method. Observing each method's abundance maps separately, without considering the scaling factors, CLSU and SUnSAL encounter similar troubles, where the abundances generally exceed 1 as shown in the second and fourth rows of Fig. \ref{fig:AbundancesUrban}, which leads to difficulty distinguishing the abundance maps of CLSU and SUnSAL. Once the scaling factors are considered, SCLSU immediately shows a competitive result, although it still cannot match SSUnSAL, especially in the identification of asphalt and grass. Pure material identification and clear abundance maps are the more reasonable and desirable results given by our model.
\begin{figure}[!t]
	  \centering
		\subfigure[A false color image]{
			\includegraphics[width=2.5cm,height=3cm]{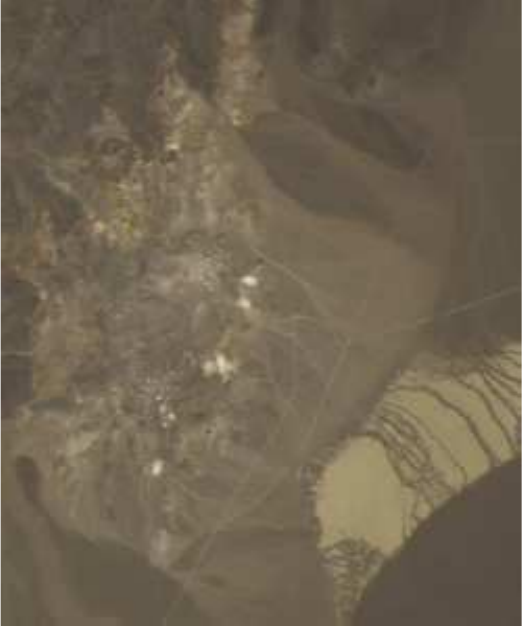}
            \label{fig:ImageCupritea}
		}
		\subfigure[Endmembers]{
			\includegraphics[width=4cm,height=3cm]{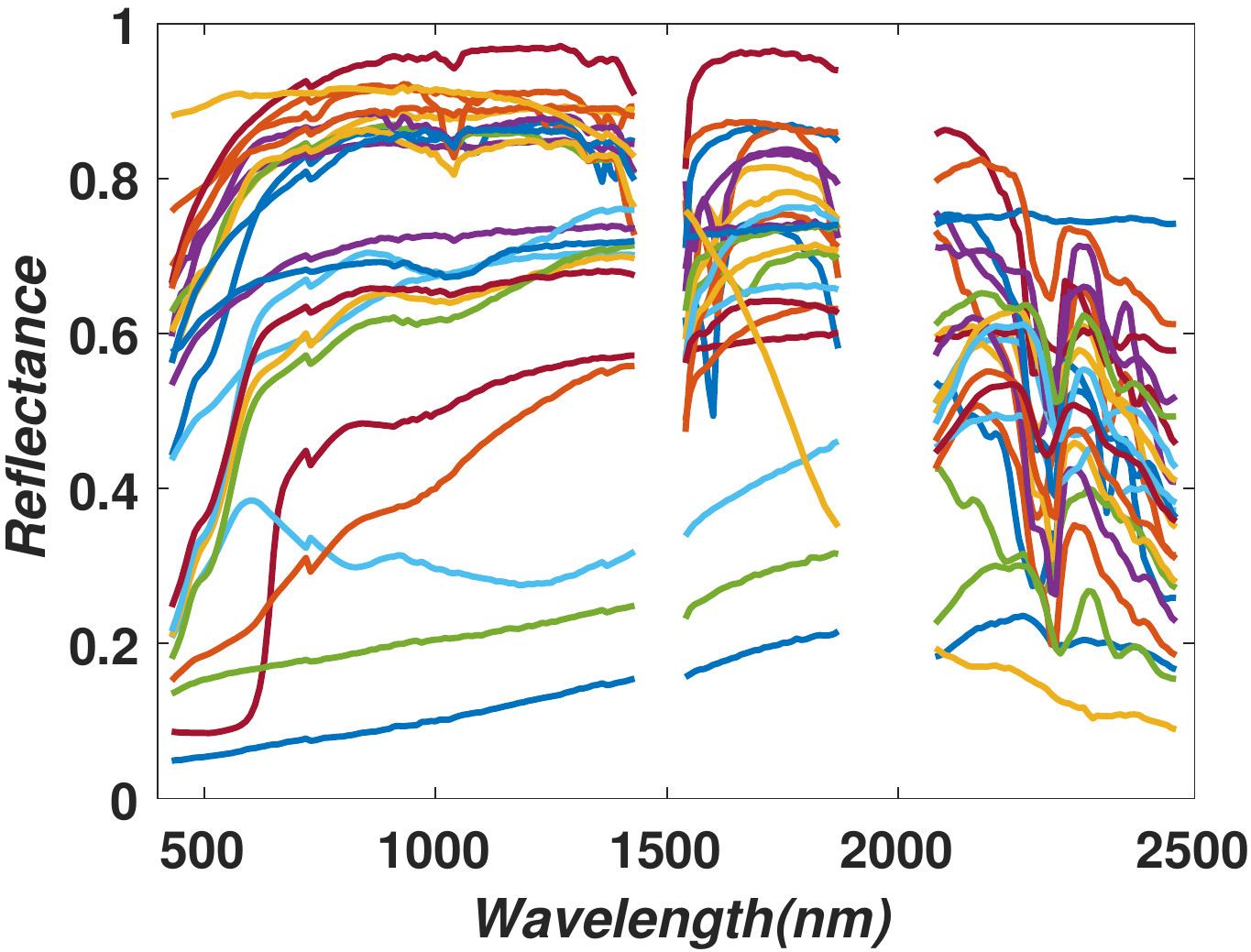}
            \label{fig:ImageCupriteb}
		}
         \caption{A false color image of the Cuprite data and the endmember dictionary constructed by spectral library.}
\label{fig:ImageCuprite}
\end{figure}
\begin{figure}[!t]
	  \centering
			\includegraphics[width=0.45\textwidth]{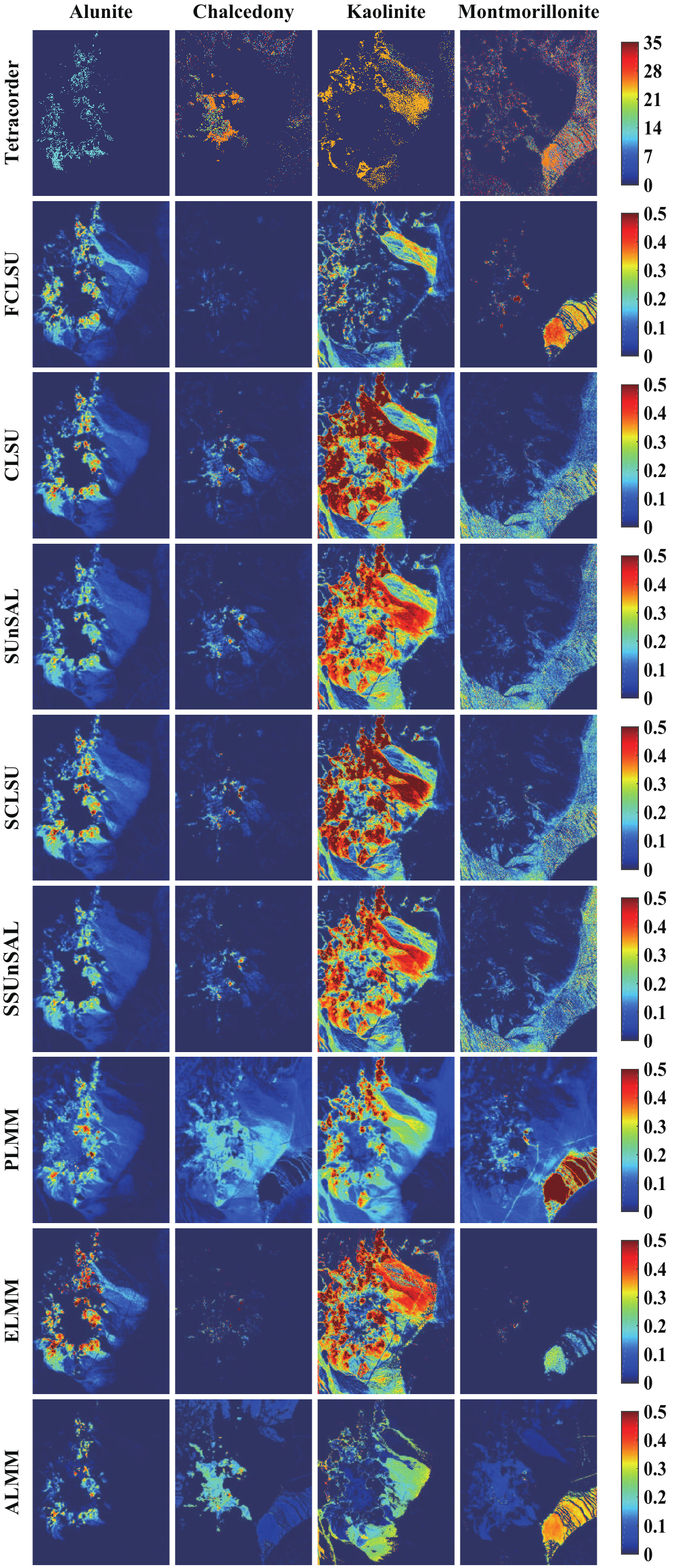}
         \caption{The abundance maps estimated by different SU methods and the first row shows the so-called ground truth generated by Tetracorder.}
\label{fig:AbundanceCuprite}
\end{figure}
\begin{table*}
\small
\begin{spacing}{1.1}
\centering
\caption{Quantitative performance comparison with the different algorithms on the Cuprite data. The best one in shown in bold.}
\smallskip
\begin{tabular}{|c|c|c|c|c|c|c|c|c|}
\hline Algorithm&FCLSU&CLSU&SUnSAL&SCLSU&SSUnSAL&PLMM&ELMM&ALMM\\
\hline OA (\%)&31.03&62.75&67.03&62.75&67.03&57.98&52.06&\bf74.17\\
\hline rRMSE&0.0470&0.0213&0.0206&0.0213&0.0206&0.367&0.0181&\bf0.0003\\
\hline aSAM&2.4588&1.9083&1.9278&1.9083&1.9278&2.1462&1.4138&\bf0.0052\\
\hline
\end{tabular}
\end{spacing}
\end{table*}
\subsection{Second Real Data (Cuprite)}
\subsubsection{Data Description}
The second real dataset is the hyperspectral image acquired by the airborne visible-infrared imaging spectrometer (AVIRIS) over the Cuprite mining district in western Nevada, USA, which is composed of various minerals. We selected a sub-image composed of 304 $\times$ 257 pixels at a GSD of 20 m to evaluate the performance between the proposed method and the compared methods. The wavelength of 224 spectral bands ranges from 400 nm to 2500 nm with 10 nm spectral resolution. Before unmixing, bands eroded by water absorption, atmospheric effects, and noise (bands$1$-$2$, $104$-$113$, $148$-$167$, $221$-$224$) were removed; $188$ bands were used in the experiment. A false-color image of the Cuprite data is shown in Fig. \ref{fig:ImageCupritea}.
\subsubsection{Experimental Setup}
With the difficult challenges generated by the highly mixed property of the minerals and the low spatial resolution of the study image, this scene is commonly used for evaluating unmixing performance. Furthermore, data-driven endmember extraction is very challenging due to highly mixed effects. Therefore, we used the USGS spectral library to construct the endmember dictionary. The detailed procedures are as follows: First, VCA was applied for extracting $14$ endmembers. \footnote{The number of endmembers is estimated by Hysime on the subset of Cuprite.} Then, material identification was performed using the USGS spectral library and spectral feature fitting \cite{Clark1984SSF}. Next, the endmember dictionary ($\mathbf{A}$) was constructed based on identified spectral signatures from the library, whose similarity scores are higher than a threshold. \footnote{In our case, the threshold is experimentally set up as $0.93$.} Finally, $24$ spectral signatures we selected from the spectral library as the endmember dictionary, as shown in Fig. \ref{fig:ImageCupriteb}. We used the same endmember matrix for all algorithms.
\subsubsection{Results and Analysis}
Similar to the first real data, we evaluate the performance for the Cuprite data both quantitatively and visually. The classification-based evaluation and the two reconstruction indicies are summarized in Table III. There is a difference in calculating OA. Owing to highly mixed effects of the minerals, it is quite difficult to exactly estimate the number of endmembers. Therefore, in order to effectively use OA for quantitatively assessing the performance of the different algorithms, we only considered four principal minerals, i.e., alunite, chalcedony, kaolinite, and montmorillonite.

The estimated abundance maps of the four minerals are shown in Fig. \ref{fig:AbundanceCuprite}. The first row represents the reference classification maps generated by Tetracorder software \cite{Clark2003Tetracorder}. Since FCLSU fails to take spectral variability into account and while strictly following the ANC and the ASC to the abundance maps, it yielded the unexpected result of the absence of certain material, as shown in the abundance map of the material {\it Chalcedony} of Fig. \ref{fig:AbundanceCuprite}. Although the CLSU and SUnSAL algorithms can improve the visual effects by relaxing the ASC, particularly for the materials of kaolinite and montmorillonite, the range of abundances is obviously over $1$, which makes no sense in reality. With the consideration of scaling factors, scaled versions of CLSU and SUnSAL effectively show the abundances to be in the understandable range. On the other hand, the reasonable assumption that the mixed spectral signature is sparsely represented by the endmember dictionary leads to a good visual result that approaches that of Tetracorder, as shown in the comparisons between CLSU and SUnSAL as well as their scaled versions.

PLMM and ELMM tend to specify the spectral variability. Considering the spectral variability as the perturbation information, PLMM is relatively hard to detect the pure area, since the main spectral variability (scaling factors) is ignored in this model. The estimated abundance maps of PLMM in Fig. \ref{fig:AbundanceCuprite} gives consistent results. While ELMM gives one scaling factor for each endmember, which yields much clearer results. The proposed method shows the best visual resemblance, compared with the results from the Tetracorder. The abundance maps generated by the proposed ALMM are more distinct and show greater contrast, and the distribution of each material is regional as well, which implies that various spectral variabilities could be learned effectively. Consistent with the analysis above, Table III gives a similar quantitative evaluation.

\section{Conclusion}
ELMM and PLMM have their respective drawbacks. ELMM ignores those spectral variabilities that cannot be explained only by scaling factors, and it is hard to obtain a good scaling estimation due to ELMM's non-convexity. With PLMM, the perturbation information is too general to model various spectral variabilities. To this end, we proposed a novel spectral mixture model, called ALMM, which considers not only the principal scaling factor but also other various spectral variabilities by introducing the spectral variability dictionary to expand the scalability of the endmember dictionary. To effectively promote spectral unmixing based on the proposed method, we modeled the spectral variability as low-coherent with the endmember dictionary and developed an algorithm for learning the spectral variability dictionary. By analyzing experimental results on a synthetic dataset and two real datasets, we found that the methods taking the spectral variability into consideration are generally superior to those that do not. More notably, the proposed method is able to obtain a more accurate abundance estimation compared to other state-of-the-art algorithms, since we separately model the spectral variability as scaling factors and other spectral variability according to their distinctive properties.

\appendices
\section{Solution to ALMM-based Spectral Unmixing}
The object function in Eq. (\ref{eq20}) is not convex with respect to all variables simultaneously, but it is a convex problem regarding the separate variable when other variables are fixed. As a result, we successively minimize $\mathscr{L}_{U}$ with respect to $\mathbf{x}_{k},S_{k},\mathbf{b}_{k},\mathbf{g}_{k},\mathbf{h}_{k},\pmb{\lambda}_{k}, \pmb{\nu}_{k}$ as follows:

{\it Optimization with respect to $\mathbf{x}_{k}$ and $S_{k}$:} Bundle $S_{k}\mathbf{A}$ as $\mathbf{D}$ and this subproblem can be written as
\begin{equation}
\label{eq25}
\begin{aligned}
       \mathop{arg \min}_{\mathbf{x}_{k}}&\frac{1}{2}\norm{\mathbf{y}_{k}-\mathbf{D}\mathbf{x}_{k}-\mathbf{E}\mathbf{b}_{k}}_{2}^{2}+\pmb{\lambda}_{k}^{\T}(\mathbf{g}_{k}-\mathbf{x}_{k})\\
       &+\pmb{\nu}_{k}^{\T}(\mathbf{h}_{k}-\mathbf{x}_{k})+\frac{\mu}{2}\norm{\mathbf{g}_{k}-\mathbf{x}_{k}}_{2}^{2}+\frac{\mu}{2}\norm{\mathbf{h}_{k}-\mathbf{x}_{k}}_{2}^{2},
\end{aligned}
\end{equation}
which has a closed-form solution:
\begin{equation}
\label{eq26}
\begin{aligned}
       \mathbf{x}_{k}\leftarrow&(\mathbf{D}^{\T}\mathbf{D}+2\mu\mathbf{I})^{-1}\\
       &\times (\mu \mathbf{g}_{k}+\pmb{\lambda}_{k}+\mu \mathbf{h}_{k}+\pmb{\nu}_{k}+ \mathbf{D}^{\T}\mathbf{y}_{k}-\mathbf{D}^{\T}\mathbf{E}\mathbf{b}_{k}).
\end{aligned}
\end{equation}
Inspired by SCLSU, we further update $\mathbf{x}_{k}$ in order to satisfy the sum-to-one constraint by
\begin{equation}
\label{eq27}
\begin{aligned}
     \mathbf{x}_{k}\leftarrow\mathbf{x}_{k}/\mathbf{1}^{\T}\mathbf{x}_{k}.
\end{aligned}
\end{equation}
Hereinafter, bundle $\mathbf{A}\mathbf{x}_{k}$ as Z and $S_{k}$ can be estimated by solving NNLS problem \cite{Kim2010NNLS}:
\begin{equation}
\label{eq28}
\begin{aligned}
       \hat{S}_{k}=\mathop{arg \min}_{S_{k}\succeq 0}\frac{1}{2}\norm{(\mathbf{y}_{k}-\mathbf{E}\mathbf{b}_{k})-S_{k}\mathbf{Z}}_{2}^{2}.
\end{aligned}
\end{equation}

{\it Optimization with respect to $\mathbf{b}_{k}$:} For $\mathbf{b}_{k}$, the optimization problem is
\begin{equation}
\label{eq29}
\begin{aligned}
      \mathop{arg \min}_{\mathbf{b}_{k}} \frac{1}{2}\norm{\mathbf{y}_{k}-(S_{k}\mathbf{A})\mathbf{x}_{k}-\mathbf{E}\mathbf{b}_{k}}_{2}^{2}+\frac{\beta}{2} \norm{\mathbf{b}_{k}}_{2}^{2},
\end{aligned}
\end{equation}
which is readily solved by
\begin{equation}
\label{eq30}
\begin{aligned}
      \mathbf{b}_{k}\leftarrow (\mathbf{E}^{\T}\mathbf{E}+\beta\mathbf{I})^{-1}(\mathbf{E}^{\T}\mathbf{y}_{k}-S_{k}\mathbf{E}^{\T}\mathbf{A}\mathbf{x}_{k}).
\end{aligned}
\end{equation}

{\it Optimization with respect to $\mathbf{g}_{k}$:} The subproblem of $\mathbf{g}_{k}$ can be written as
\begin{equation}
\label{eq31}
\begin{aligned}
      \mathop{arg \min}_{\mathbf{g}_{k}}\frac{\alpha}{2}\norm{\mathbf{g}_{k}}_{1} +\pmb{\lambda}_{k}^{\T}(\mathbf{g}_{k}-\mathbf{x}_{k})+\frac{\mu}{2}\norm{\mathbf{g}_{k}-\mathbf{x}_{k}}_{2}^{2},
\end{aligned}
\end{equation}
whose solution is the well-known {\it soft threshold} \cite{BDias3}:
\begin{equation}
\label{eq32}
\begin{aligned}
       \mathbf{g}_{k}\leftarrow \max \{\mathbf{0},\norm{\mathbf{x}_{k}-\pmb{\lambda}/\mu}_{1}-\alpha/\mu \}sign(\mathbf{x}_{k}-\pmb{\lambda}/\mu),
\end{aligned}
\end{equation}
where $sign(\bullet)$ is defined by
\begin{equation}
\label{eq33}
\begin{aligned}
       sign(\bullet)=
       \begin{cases}
       \hspace{0.15cm}1\hspace{0.12cm}, & \bullet \succeq 0\\ -1, & \bullet \prec 0.
       \end{cases}
\end{aligned}
\end{equation}

{\it Optimization with respect to $\mathbf{h}_{k}$:} The optimization problem of $\mathbf{h}_{k}$ is
\begin{equation}
\label{eq34}
\begin{aligned}
      \mathop{arg \min}_{\mathbf{h}_{k}}\pmb{\nu}_{k}^{\T}(\mathbf{h}_{k}-\mathbf{x}_{k})+\frac{\mu}{2}\norm{\mathbf{h}_{k}-\mathbf{x}_{k}}_{2}^{2}+l_{R}^{+}(\mathbf{h}_{k}).
\end{aligned}
\end{equation}
Here the update rule for $\mathbf{h}_{k}$ is
\begin{equation}
\label{eq35}
\begin{aligned}
       \mathbf{h}_{k}\leftarrow \max \{\mathbf{0},\mathbf{x}_{k}-\pmb{\nu}/\mu \}.
\end{aligned}
\end{equation}

{\it Lagrange multipliers update $\pmb{\lambda}_{k}$ and $\pmb{\nu}_{k}$:} Before stepping into the next iteration, Lagrange multipliers need to be updated by
\begin{equation}
\label{eq36}
\begin{aligned}
       \pmb{\lambda}_{k}\leftarrow\pmb{\lambda}_{k}+\mu (\mathbf{g}_{k}-\mathbf{x}_{k}),\quad \pmb{\nu}_{k}\leftarrow\pmb{\nu}_{k}+\mu (\mathbf{h}_{k}-\mathbf{x}_{k}).
\end{aligned}
\end{equation}

\section{Solution to ALMM-based Spectral Variability Dictionary Learning}
To solve Eq. (\ref{eq21}), we have:

{\it Optimization with respect to $\mathbf{M}$:} The optimization problem can be formulated as follows:
\begin{equation}
\label{eq37}
\begin{aligned}
      \mathop{arg \min}_{\mathbf{M}} &\frac{1}{2}\norm{\mathbf{Y}-\mathbf{A}\mathbf{M}-\mathbf{E}\mathbf{B}}_{\F}^{2}+\pmb{\Omega}^{\T}(\mathbf{M}-\mathbf{XS})\\
      &+\frac{\xi}{2}\norm{\mathbf{M}-\mathbf{XS}}_{\F}^{2},
\end{aligned}
\end{equation}
which can be quickly solved by
\begin{equation}
\label{eq38}
\begin{aligned}
       \mathbf{M}\leftarrow (\mathbf{A}^{\T}\mathbf{A}+\xi\mathbf{I})^{-1}(\mathbf{A}^{\T}\mathbf{Y}-\mathbf{A}^{\T}\mathbf{E}\mathbf{B}+\xi\mathbf{XS}-\pmb{\Omega}).
\end{aligned}
\end{equation}

{\it Optimization with respect to $\mathbf{B}$:} The analytical solution for $\mathbf{B}$ can be simply obtained by the matrix form of Eq. (\ref{eq30})
\begin{equation}
\label{eq39}
\begin{aligned}
       \mathbf{B}\leftarrow (\mathbf{E}^{\T}\mathbf{E}+\beta\mathbf{I})^{-1}(\mathbf{E}^{\T}\mathbf{Y}-\mathbf{E}^{\T}\mathbf{A}\mathbf{M}).
\end{aligned}
\end{equation}

{\it Optimization with respect to $\mathbf{X}$:} The optimization problem is expressed as follows:
\begin{equation}
\label{eq40}
\begin{aligned}
      \mathop{arg \min}_{\mathbf{X}} \; & \pmb{\Lambda}^{\T}(\mathbf{G}-\mathbf{X})+\pmb{V}^{\T}(\mathbf{H}-\mathbf{X})+\pmb{\Omega}^{\T}(\mathbf{M}-\mathbf{XS})\\
      &+\frac{\xi}{2}\norm{\mathbf{G}-\mathbf{X}}_{\F}^{2}+\frac{\xi}{2}\norm{\mathbf{H}-\mathbf{X}}_{\F}^{2}+\frac{\xi}{2}\norm{\mathbf{M}-\mathbf{XS}}_{\F}^{2}.
\end{aligned}
\end{equation}
The solution of Eq. (\ref{eq40}) is given by
\begin{equation}
\label{eq41}
\begin{aligned}
       \mathbf{X}\leftarrow &(\xi\mathbf{G}+\pmb{\Lambda}+\xi\mathbf{H}+\pmb{V}+\pmb{\Omega}\mathbf{S}^{\T}+\xi\mathbf{M}\mathbf{S}^{\T})\\
       &\times (\xi \mathbf{S}\mathbf{S}^{\T}+2\xi\mathbf{I})^{-1}.
\end{aligned}
\end{equation}
In order to remove the scaling factors while satisfying the sum-to-one constraint, $\mathbf{X}$ is rewritten as a matrix form of Eq. (\ref{eq27}).
\begin{equation}
\label{eq42}
\begin{aligned}
       \mathbf{X} \leftarrow\mathbf{X}\oslash (\mathbf{1}^{\T}\mathbf{X}),
\end{aligned}
\end{equation}
where $\oslash$ is defined as a term-wise Hadamard division.

{\it Optimization with respect to $\mathbf{S}$:} Subsequently, the variable $\mathbf{S}$ can be collected by solving the following problem:
\begin{equation}
\label{eq43}
\begin{aligned}
      \mathop{arg \min}_{\mathbf{S}} \; &\pmb{\Omega}^{\T}(\mathbf{M}-\mathbf{XS})+\pmb{\Delta}^{\T}(\mathbf{\T}-\mathbf{S})+\frac{\xi}{2}\norm{\mathbf{M}-\mathbf{XS}}_{\F}^{2}\\
      &+\frac{\xi}{2}\norm{\mathbf{T}-\mathbf{S}}_{\F}^{2},
\end{aligned}
\end{equation}
whose a closed-form solution can be obtained as
\begin{equation}
\label{eq44}
\begin{aligned}
       \mathbf{S} \leftarrow(\xi\mathbf{X}^{\T}\mathbf{X}+\xi\mathbf{I})^{-1}(\xi\mathbf{X}^{\T}\mathbf{M}+\mathbf{X}^{\T}\pmb{\Omega}+\xi\mathbf{\T}+\pmb{\Delta}).
\end{aligned}
\end{equation}

{\it Optimization with respect to $\mathbf{E}$:} The object function with respect to $\mathbf{E}$ is written as
\begin{equation}
\label{eq45}
\begin{aligned}
      \mathop{arg \min}_{\mathbf{E}}&\frac{1}{2}\norm{\mathbf{Y}-\mathbf{A}\mathbf{M}-\mathbf{E}\mathbf{B}}_{\F}^{2}+\pmb{\Pi}^{\T}(\mathbf{Q}-\mathbf{E})\\
      &+\frac{\xi}{2}\norm{\mathbf{Q}-\mathbf{E}}_{\F}^{2},
\end{aligned}
\end{equation}
which has the analytical solution of
\begin{equation}
\label{eq46}
\begin{aligned}
       &\mathbf{E}\leftarrow (\mathbf{Y}\mathbf{B}^{\T}-\mathbf{AM}\mathbf{B}^{\T}+\xi\mathbf{Q}+\pmb{\Pi})(\mathbf{B}\mathbf{B}^{\T}+\xi\mathbf{I})^{-1}.
\end{aligned}
\end{equation}

{\it Optimization with respect to $\mathbf{Q}$:} Inspired by \cite{Barchiesi2013Incoherent}, the optimization problem with the Gram matrix $\mathbf{E}^{T}\mathbf{E}$ can be effectively solved as follows: we define $\mathbf{Q}_{p}$ to be $\mathbf{Q}$ of the former step, so it can be regarded as a known matrix in the current step, and therefore the optimization problem for $\mathbf{Q}$ can be specified as
\begin{equation}
\label{eq47}
\begin{aligned}
      \mathop{arg \min}_{\mathbf{Q}} &\frac{\gamma }{2}\norm{\mathbf{A}^{\T}\mathbf{Q}}_{\F}^{2}+\frac{\eta}{2}\norm{\mathbf{Q}_{p}^{\T}\mathbf{Q}-\mathbf{I}}_{\F}^{2}+\pmb{\Pi}^{\T}(\mathbf{Q}-\mathbf{E})\\
     &+\frac{\xi}{2}\norm{\mathbf{Q}-\mathbf{E}}_{\F}^{2},
\end{aligned}
\end{equation}
which can be easily deduced as
\begin{equation}
\label{eq48}
\begin{aligned}
       &\mathbf{Q}\leftarrow (\gamma \mathbf{A}\mathbf{A}^{\T}+ \eta \mathbf{Q}_{p}\mathbf{Q}_{p}^{\T}+\xi\mathbf{I})^{-1}(\eta \mathbf{Q}_{p}+\xi\mathbf{E}-\pmb{\Pi}).
\end{aligned}
\end{equation}

{\it Optimization with respect to $\mathbf{G}$ and $\mathbf{H}$:} The two variables can be summarized using the matrix form of Eq. (\ref{eq32}) and Eq. (\ref{eq35}) as
\begin{equation}
\label{eq49}
\begin{aligned}
       \mathbf{G}\leftarrow \max \{\mathbf{0},\norm{\mathbf{X}-\pmb{\Lambda}/\xi}_{1,1}-\alpha/\xi \}sign(\mathbf{X}-\pmb{\Lambda}/\xi),
\end{aligned}
\end{equation}
\begin{equation}
\label{eq50}
\begin{aligned}
       \mathbf{H}\leftarrow\max \{\mathbf{0},\mathbf{X}-\pmb{V}/\xi \}.
\end{aligned}
\end{equation}

{\it Optimization with respect to $\mathbf{T}$:} The variable $\mathbf{T}$ can be updated by using the same rule as with $\mathbf{H}$:
\begin{equation}
\label{eq51}
\begin{aligned}
       \mathbf{T}\leftarrow\max \{\mathbf{0},\mathbf{S}-\pmb{\Delta}/\xi \}.
\end{aligned}
\end{equation}

{\it Lagrange multipliers update $\pmb{\Lambda}$, $\pmb{V}$, $\pmb{\Omega}$, $\pmb{\Pi}$ and $\pmb{\Delta}$:} Following the rule of Eq. (\ref{eq36}), these Lagrange multipliers can be updated in each iteration:
\begin{equation}
\label{eq52}
\begin{aligned}
       &\pmb{\Lambda}\leftarrow\pmb{\Lambda}+\xi (\mathbf{G}-\mathbf{X}), \quad \pmb{V}\leftarrow\pmb{V}+\xi (\mathbf{H}-\mathbf{X}),\\
       &\pmb{\Delta}\leftarrow\pmb{\Delta}+\xi (\mathbf{T}-\mathbf{S}), \quad \pmb{\Pi}\leftarrow\pmb{\Pi}+\xi (\mathbf{Q}-\mathbf{E}),\\
       &\qquad \qquad \pmb{\Omega}\leftarrow\pmb{\Omega}+\xi (\mathbf{M}-\mathbf{XS}).
\end{aligned}
\end{equation}


\section*{Acknowledgments}

The authors would like to thank Pierre-Antoine Thouvenin from IRIT (Institut de Recherche en Informatique de Toulouse) for providing the PLMM code tested in our experiments, and the Hyperspectral Digital Imagery Collection Experiment (HYDICE) for sharing the urban dataset free of charge.

\bibliographystyle{IEEEbib}
\bibliography{HDF_ref}

\begin{IEEEbiography}[{\includegraphics[width=1in,height=1.25in,clip,keepaspectratio]{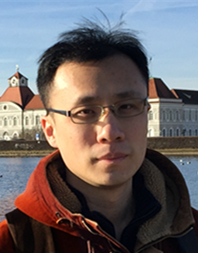}}]{Danfeng Hong}
(S'16) received the B.Sc. degree in computer science and technology from the
Neusoft College of Information, Northeastern University, China, in 2012, the M.Sc. degree in computer vision, Qingdao University, China, in 2015. He is currently pursuing his Ph.D. degree in Signal Processing in Earth Observation, Technical University of Munich (TUM), Munich Germany, and the Remote Sensing Technology Institute, German Aerospace Center (DLR), Oberpfaffenhofen, Germany.

His research interests include signal/image processing and analysis, pattern recognition, machine/deep learning and their applications in Earth Vision.
\end{IEEEbiography}
\vskip -2\baselineskip plus -1fil
\begin{IEEEbiography}[{\includegraphics[width=1in,height=1.25in,clip,keepaspectratio]{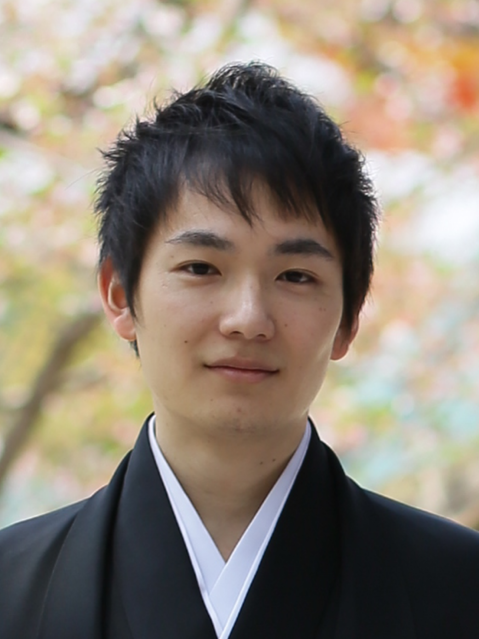}}]{Naoto Yokoya}
Yokoya (S'10–M'13) received the M.Sc. and Ph.D. degrees in aerospace engineering from the University of Tokyo, Tokyo, Japan, in 2010 and 2013, respectively.

From 2012 to 2013, he was a Research Fellow with Japan Society for the Promotion of Science, Tokyo, Japan. From 2013 to 2017, he was an Assistant Professor with the University of Tokyo In 2015-2017, he was also an Alexander von Humboldt Research Fellow with the German Aerospace Center (DLR), Oberpfaffenhofen, and Technical University of Munich (TUM), Munich, Germany. Since 2018, he leads the Geoinformatics Unit at the RIKEN Center for Advanced Intelligence Project (AIP), RIKEN, Tokyo, Japan. His research interests include image analysis and data fusion in remote sensing.

Since 2017, he is a Co-chair of IEEE Geoscience and Remote Sensing Image Analysis and Data Fusion Technical Committee. In 2017, he won the Data Fusion Contest 2017 organized by the Image Analysis and Data Fusion Technical Committee of the IEEE Geoscience and Remote Sensing Society. His model was the most accurate among over 800 submissions.
\end{IEEEbiography}

\begin{IEEEbiography}[{\includegraphics[width=1in,height=1.25in,clip,keepaspectratio]{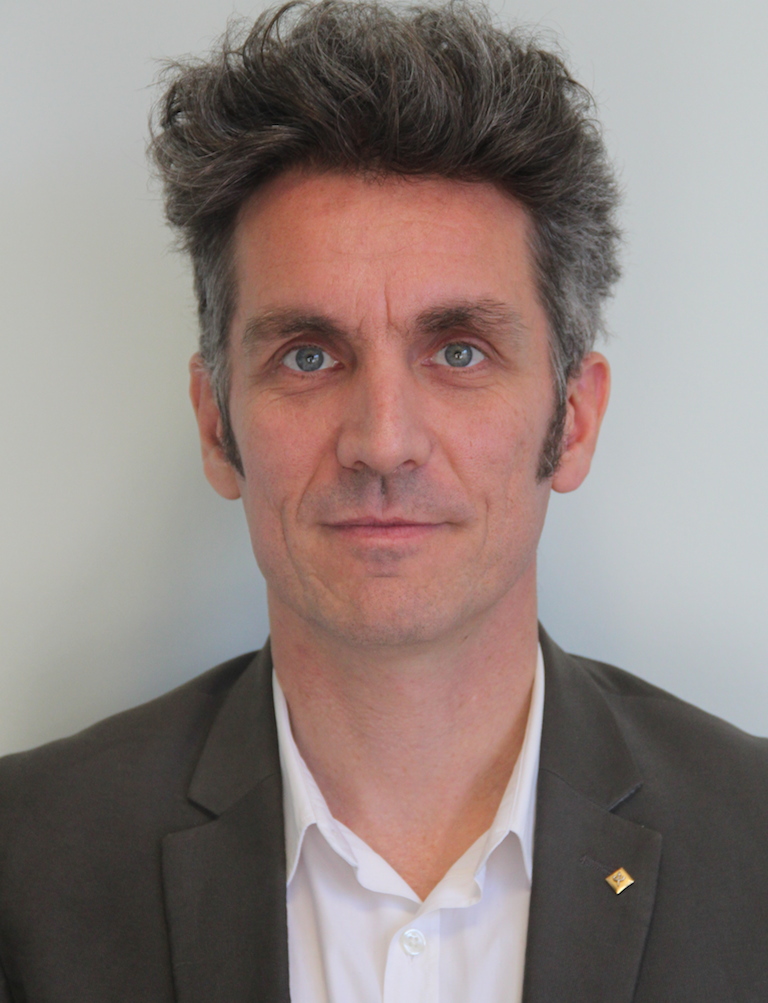}}]{Jocelyn Chanussot}
(M'04-–SM'04–-F'12) received the M.Sc. degree in electrical engineering from the Grenoble Institute of Technology (Grenoble INP), Grenoble, France, in 1995, and the Ph.D. degree from the Université de Savoie, Annecy, France, in 1998. In 1999, he was with the Geography Imagery Perception Laboratory for the Delegation Generale de l'Armement (DGA - French National Defense Department). Since 1999, he has been with Grenoble INP, where he is currently a Professor of signal and image processing. He is conducting his research at GIPSA-Lab. His research interests include image analysis, multicomponent image processing, nonlinear filtering, and data fusion in remote sensing. He has been a visiting scholar at Stanford University, KTH (Sweden) and NUS (Singapore). Since 2013, he is an Adjunct Professor of the University of Iceland. In 2015-2017, he was a visiting professor at the University of California, Los Angeles (UCLA).

   Dr. Chanussot is the founding President of IEEE Geoscience and Remote Sensing French chapter (2007-2010) which received the 2010 IEEE GRSS Chapter Excellence Award. He was the General Chair of the first IEEE GRSS Workshop on Hyperspectral Image and Signal Processing, Evolution in Remote sensing (WHISPERS). He was the Chair (2009-2011) and Co-chair of the GRS Data Fusion Technical Committee (2005-2008). He was a member of the Machine Learning for Signal Processing Technical Committee of the IEEE Signal Processing Society (2006-2008) and the Program Chair of the IEEE International Workshop on Machine Learning for Signal Processing, (2009). He was an Associate Editor for the IEEE Geoscience and Remote Sensing Letters (2005-2007) and for Pattern Recognition (2006-2008). He was the Editor-in-Chief of the IEEE Journal of Selected Topics in Applied Earth Observations and Remote Sensing (2011-2015). Since 2007, he is an Associate Editor for the IEEE Transactions on Geoscience and Remote Sensing, and since 2018, he is also an Associate Editor for the IEEE Transactions on Image Processing. In 2013, he was a Guest Editor for the Proceedings of the IEEE and in 2014 a Guest Editor for the IEEE Signal Processing Magazine. He is a Fellow of the IEEE and a member of the Institut Universitaire de France (2012-2017).
\end{IEEEbiography}
\vskip -2\baselineskip plus -1fil
\begin{IEEEbiography}[{\includegraphics[width=1in,height=1.25in,clip,keepaspectratio]{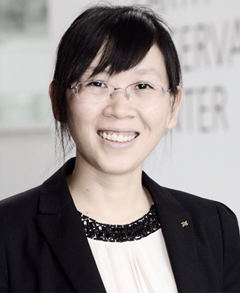}}]{Xiao Xiang Zhu}(S'10--M'12--SM'14) received the Master (M.Sc.) degree, her doctor of engineering (Dr.-Ing.) degree and her “Habilitation” in the field of signal processing from Technical University of Munich (TUM), Munich, Germany, in 2008, 2011 and 2013, respectively.

She is currently the Professor for Signal Processing in Earth Observation (www.sipeo.bgu.tum.de) at Technical University of Munich (TUM) and German Aerospace Center (DLR); the head of the department ``EO Data Science'' at DLR's Earth Observation Center; and the head of the Helmholtz Young Investigator Group ``SiPEO'' at DLR and TUM. Prof. Zhu was a guest scientist or visiting professor at the Italian National Research Council (CNR-IREA), Naples, Italy, Fudan University, Shanghai, China, the University  of Tokyo, Tokyo, Japan and University of California, Los Angeles, United States in 2009, 2014, 2015 and 2016, respectively. Her main research interests are
  remote sensing and Earth observation, signal processing, machine learning and data science, with a special application focus on global urban mapping.

  Dr. Zhu is a member of young academy (Junge Akademie/Junges Kolleg) at the Berlin-Brandenburg Academy of Sciences and Humanities and the German National  Academy of Sciences Leopoldina and the Bavarian Academy of Sciences and Humanities. She is an associate Editor of IEEE Transactions on Geoscience and Remote Sensing.
\end{IEEEbiography}

\end{document}